\documentclass[sigconf]{acmart}
\usepackage{colortbl}
\usepackage{graphicx}
\usepackage{soul}
\usepackage{enumitem}
\usepackage{cleveref}
\usepackage{csquotes}
\usepackage{multirow}
\usepackage{longtable}
\usepackage{array}
\usepackage{ragged2e}

\newcommand{\cspn}{Jane Foster}

\newcommand{\rev}[1]{{\color{black} #1}}

\AtBeginDocument{%
  \providecommand\BibTeX{{%
    \normalfont B\kern-0.5em{\scshape i\kern-0.25em b}\kern-0.8em\TeX}}}

\makeatletter
\newdimen\SOUL@dimen 
\def\SOUL@ulunderline#1{{%
    \setbox\z@\hbox{#1}%
    \SOUL@dimen=\wd\z@ 
    \dimen@i=\SOUL@uloverlap
    \advance\SOUL@dimen2\dimen@i 
    \rlap{%
        \null
        \kern-\dimen@i
        \SOUL@ulcolor{\SOUL@ulleaders\hskip\SOUL@dimen}
    }%
    \unhcopy\z@
}}

\makeatother

\setcopyright{acmlicensed}
\copyrightyear{2025}
\acmYear{2025}
\acmDOI{3708359.3712110}

\acmConference[IUI '25]{Make sure to enter the correct
  conference title from your rights confirmation emai}{March 24-27, 2025}{Cagliari, Italy}
\acmBooktitle{IUI 2025: The 30th Annual ACM Conference on Intelligent User Interfaces,
 March 24--27, 2025, Cagliari, Italy} 
\acmISBN{979-8-4007-1306-4/25/03}

\begin{document}

\title{Words as Bridges: Exploring Computational Support for Cross-Disciplinary Translation Work}

\author{Calvin Bao}
\email{csbao@umd.edu}
\orcid{1234-5678-9012} 
\affiliation{%
  \institution{University of Maryland, College Park}
  \streetaddress{8125 Paint Branch Drive}
  \city{College Park}
  \state{Maryland}
  \country{USA}
  \postcode{20742}
}

\author{Yow-Ting Shiue}
\email{orina.shiue@aif.tw}
\affiliation{%
  \institution{University of Maryland, College Park}
  \streetaddress{8125 Paint Branch Drive}
  \city{College Park}
  \state{Maryland}
  \country{USA}
  \postcode{20742}
}
\author{Marine Carpuat}
\email{marine@umd.edu}
\affiliation{%
  \institution{University of Maryland, College Park}
  \streetaddress{8125 Paint Branch Drive}
  \city{College Park}
  \state{Maryland}
  \country{USA}
  \postcode{20742}
}

\author{Joel Chan}
\email{joelchan@umd.edu}
\affiliation{
  \institution{University of Maryland, College Park}
  \streetaddress{8125 Paint Branch Drive}
  \city{College Park}
  \state{Maryland}
  \country{USA}
  \postcode{20742}
}

\renewcommand{\shortauthors}{Bao et al.}

\onecolumn
\begin{abstract}

Scholars often explore literature outside of their home community of study. This exploration process is frequently hampered by field-specific jargon. Past computational work often focuses on supporting translation work by removing jargon through simplification and summarization; here, we explore a different approach that preserves jargon as useful bridges to new conceptual spaces. Specifically, we cast different scholarly domains as different language-using communities, and explore how to adapt techniques from unsupervised cross-lingual alignment of word embeddings to explore conceptual alignments between domain-specific word embedding spaces.We developed a prototype cross-domain search engine that uses aligned domain-specific embeddings to support conceptual exploration, and tested this prototype in two case studies. We discuss qualitative insights into the promises and pitfalls of this approach to translation work, and suggest design insights for future interfaces that provide computational support for cross-domain information seeking.

\end{abstract}



\maketitle

\section{Introduction}
\label{sec:intro}


Scholars often explore literature outside of their home community of study.
When crossing domains or disciplines, jargon can be a major challenge \citep{lucyWordsGatekeepersMeasuring2023,palmerInformationWorkInterdisciplinary2002,palmerWorkBoundariesScience2001}.
Researchers may struggle to formulate effective queries to search for papers related to a specific concept they have in mind, because the same or similar concept may be referred to with very different terms. 
For instance, while research on creativity in Psychology talks about ``examples'' and ``structured imagination'', researchers in Engineering may talk about similar things in terms of ``inspirations'' and ``precedents''. The same surface form may also map to very different underlying concepts: for instance, the term ``stimulus'' in Psychology often refers to stimuli used in experiments, or that are perceived by participants, which is a distinct concept from usage in the Management and Organization science literature: specific fiscal interventions to induce more economic activity. To navigate these terminology barriers and build understandings of new domains, researchers must do difficult and time-consuming ``translation work"  \citep{palmerInformationWorkInterdisciplinary2002}.


Some research has approached addressing these terminology barriers as a text-simplification problem, exploring techniques such as text summarization/simplification to \textit{remove} jargon. For instance, \cite{augustPaperPlainMaking2023} generates plain language summaries and key questions to help lay readers understand medical research papers. While simplification is an appropriate way to make research more accessible to lay users, we worry that it might hamper the creative potential of juxtaposing the nuances of specialized concepts from a different field with concepts you may already be familiar with. Other approaches involve finding analogical relationships between papers, focusing on overlaps in problems or solutions between authors \cite{portenoyBurstingScientificFilter2022} or research papers \cite{kangAugmentingScientificCreativity2022,chanSOLVENTMixedInitiative2018}. This approach is well-suited for cross-domain \textit{problem solving}, but may be less suited for more conceptual or theoretical explorations.


In this paper, we frame the term mapping problem in scholarly cross-domain information seeking as a \textit{translation} problem, taking our cue from research on translation work in interdisciplinary research \cite{palmerInformationWorkInterdisciplinary2002}. Rather than removing jargon, the goal in this framing is to identify novel and fruitful conceptual alignments between specialized concepts that might lead to novel insights. With this framing, we investigate how to adapt Natural Language Processing (NLP) techniques that seek to align distinct conceptual spaces, whether across languages, or between distinct time periods or subcommunities using the same language.
For instance, multilingual NLP techniques such as machine translation have been used to not only translate between two very different languages, but also quite similar ones such as dialects \citep{Wan2019UnsupervisedND}.
It may also be applied to ``translation'' within the same language but diachronically (e.g. ancient to modern Chinese, \cite{chang-etal-2021-time}) or synchronically (e.g. across partisan divides \cite{KhudaBukhsh2020WeDS}).
We hypothesize that we might similarly be able to treat different research communities as speaking different languages and apply multilingual NLP techniques to assist with translating concepts between scholarly domains.


As a first step, we explore how to adapt techniques from unsupervised cross-lingual alignment of word embeddings to explore conceptual alignments between domain-specific word embedding spaces. 
This family of techniques aligns a source monolingual embedding space to a target monolingual embedding space. This alignment then allows for computing word similarities across the two embedding spaces. Here, we build ``monolingual'' embedding spaces for individual research communities, and then construct computational alignments between the communities. We explore two well-known techniques for cross-lingual embedding alignment --- MUSE \cite{Conneau2017WordTW}, and VecMap \cite{artetxe-etal-2018-robust}. 


To investigate the promise and challenges of this approach, we developed a prototype cross-domain search engine that uses aligned research-domain-specific embeddings to support conceptual exploration, and tested this prototype across two case studies with PhD-level interdisciplinary scholars exploring concepts across scholarly domains. In these case studies, cross-lingual alignments returned more novel and relevant alignments of concepts between domains compared to monolingual baselines, and comparable performance to a strong LLM baseline. 

These results suggest that the analogy between multilingual NLP and cross-disciplinary translation work is fruitful and worth exploring further. We summarize computational and design implications from our design iterations and case studies for future interfaces that provide computational support for scholarly cross-domain information seeking.




In sum, in this paper, we contribute:
\begin{itemize}
    \item \textbf{A novel computational approach for supporting cross-domain translation work}, based on an adaptation of unsupervised cross-lingual alignment of word embeddings from multilingual NLP
    \item \textbf{A prototype cross-domain search engine} that supports cross-domain translation work using cross-lingual alignment techniques
    \item \textbf{Empirical proof-of-concept from two case studies} with PhD-level interdisciplinary scholars that validate the promise of adapting multilingual NLP techniques for cross-domain translation work, and suggest future directions for developing this solution approach
\end{itemize}

\rev{We make our code and data publicly available.\footnote{https://github.com/oasisresearchlab/crossdisciplinary-embeddings}} 

\section{Related Work}

Our work advances research on intelligent systems for cross-domain and cross-disciplinary scholarly information seeking by conceptualizing cross-disciplinary information seeking as translation work. This conceptualization motivates us to adapt distributional semantics techniques for aligning terms across domains to support cross-domain information seeking.

\subsection{Cross-Domain Information Seeking as Translation Work}

Cross-domain information seeking has been extensively studied in the field of library and information science \cite{palmerInformationResearchInterdisciplinarity2017}. Early work documented how information \textit{scatter} --- the fragmentation of information across sources, organizations, and disciplinary clusters --- poses substantial challenges to scholars and librarians who seek relevant information via browsing or directed search in databases \cite{batesLearningInformationSeeking1996}. The boundaries that need to be crossed to discover and integrate new ideas may be drawn at disciplinary boundaries, but are also often defined by ``invisible colleges'' \cite{craneInvisibleCollegesDiffusion1972} of scholars and/or works that cite each other \cite{sandstromScholarlyCommunicationSocioecological2001}. These ``invisible colleges'' may partially cross disciplinary boundaries but cohere around common topics of interest and ways of approaching those topics \cite{hjorlandEpistemologySociocognitivePerspective2002}, such as ethnic-racial studies. In this paper, we use the term ``cross-domain information seeking'' as an umbrella term to refer to the range of information seeking activities that cross disciplinary, sociological, paradigmatic, and other significant boundaries between research communities.

Palmer's \cite{palmerStructuresStrategiesInterdisciplinary1999,palmerWorkBoundariesScience2001} seminal studies of the information work of interdisciplinary scientists documented how scientists ``import'' relevant information from outside their specialties using a complex mix of processes and practices, such as browsing and consulting a variety of journals, cultivating networks of personal contacts across disciplines, and even embedding graduate students as intermediaries to other labs to identify and transfer tacit knowledge. Of particular interest is the framing of interdisciplinary information work as ``translation work'': drawing on actor-network theory \cite{latourReassemblingSocialIntroduction2007}, Palmer describes how 
interdisciplinary scholars ``define, interpret, and redefine new information, retaining essential elements of the original context while revising and reapplying it for their own purposes'' \cite[p. 107]{palmerInformationWorkInterdisciplinary2002}. 
In this conceptualization of translation work, jargon can be a barrier, and less specialized language can facilitate broader cross-discipline engagement with ideas \cite{lucyWordsGatekeepersMeasuring2023,mcmahanAmbiguityEngagement2018}, but jargon may also act as \textit{bridges} to novel perspectives worth importing and integrating into new insights. We see a potentially fruitful analogy here to the ways in which variations in semantic overlap between languages can signal opportunities for conceptual expansion: for instance, Dörk and colleagues \cite{dorkInformationFlaneurFresh2011} drew on the French noun \textit{flâneur} --- which refers to an urban man who wanders through the boulevards and cafes of Paris without a driving goal other than experiencing city life --- to develop a novel conceptual vision of information seeking that focuses on open-ended, enjoyable construction of explorable information spaces.

We are motivated by this conceptualization of translation work to investigate a jargon-preserving computational approach to facilitating cross-domain information seeking, potentially revealing new and undiscovered connections between scholarly domains.

\subsection{Intelligent Interactive Systems for Supporting Scholarly Information Seeking}

The field of HCI and intelligent user interfaces is home to substantial, long-running bodies of work on intelligent systems for supporting scholarly information seeking, from citation recommendation and information retrieval approaches \cite{leePaperWeaverEnrichingTopical2024,palaniRelatedlyScaffoldingLiterature2023,portenoyBurstingScientificFilter2022,chanSOLVENTMixedInitiative2018,kangAugmentingScientificCreativity2022,kangComLitteeLiteratureDiscovery2023}, to novel search and sensemaking interfaces \cite{kangThreddyInteractiveSystem2022,chauApoloMakingSense2011,rachatasumritCiteReadIntegratingLocalized2022,kangSynergiMixedInitiativeSystem2023,changCiteSeeAugmentingCitations2023,dunneRapidUnderstandingScientific2012}. Work in this area often deals with scholarly information seeking in a domain-centric or domain-agnostic manner: users are either explicitly or implicitly modeled as searching in a single search space of ``the scientific literature''. 

Our work is situated in a smaller body of work that aims to directly support information seeking that crosses domain or disciplinary boundaries. Some of this work uses citation-based approaches to identify partitions in the literature and enable users to bridge those partitions: for example, literature-based discovery systems \cite{sebastianEmergingApproachesLiteraturebased2017} aim to discover conceptual chains that bridge disjoint citation networks. In a classic motivating success story for this approach, Swanson \cite{swansonUndiscoveredPublicKnowledge1986} identified a citation network A where ``fish oil'' often co-occurs with ``blood viscosity'', and another citation network B where ``blood viscosity'' and ''Raynaud's syndrome'' often co-occur; A and B lacked citation connections between them, and Swanson proposed that there may be a novel opportunity to explore how fish oil might treat Raynaud's syndrome. Another example is Portenoy's \cite{portenoyBurstingScientificFilter2022} system that also uses citation network gaps to recommend novel \textit{authors} whose work a scholar might find inspirational. 

Another line of work develops content-based approaches to bridging domain and disciplinary gaps. Some, for example, focus on simplifying or removing jargon, often to bridge gaps between expert and lay audiences \cite{augustPaperPlainMaking2022}. Other systems draw on conceptualizations of analogical problem solving to identify \textit{analogical} relationships between research papers that focus on shared problems in spite of or in opposition to domain or topical similarity, enabling discovery of cross-domain papers that may be relevant to a research problem \cite{chanSOLVENTMixedInitiative2018,kangAugmentingScientificCreativity2022,radenskyScideatorHumanLLMScientific2024}. Finally, recent work has begun to explore the use of Large Language Models (LLMs) to support cross-disciplinary information seeking; these approaches prompt LLMs to generate research questions or queries from the perspective of other disciplines or domains, which can then be used as inspiration or queries for information seeking \cite{zhengDiscipLinkUnfoldingInterdisciplinary2024,liuPersonaFlowBoostingResearch2024}. 

Our work is most closely related to content-based approaches to supporting cross-domain information seeking. We extend this body of work by focusing on identifying novel and interesting specialized \textit{concepts} (vs. authors or papers) across domain boundaries.

\subsection{Distributional Semantics Methods for Aligning Terms across Domains}

We explore how our goal of identifying interesting conceptual alignments between scholarly domains can draw on distributional semantics methods from NLP that learn computational representations for words based on the context of how they co-occur with other words.  This process is motivated by the ``distributional hypothesis,'' \cite{Sahlgren2008TheDH} which hypothesizes that \textit{distributional similarity} is correlated with \textit{meaning similarity}, so one could be approximated by the other. The learned distributional representations of words as vectors enables the use of measures like cosine similarity between word vectors to discover related terms. 

For our purposes, building a single representation of word meaning is likely suboptimal, since it would fail to account for variations in language use across specific domains or communities. We thus draw on methods for inducing \textit{alignments} between two distinct vector spaces \cite{kulshreshtha-etal-2020-cross, marchisio-etal-2022-isovec, patra-etal-2019-bilingual, artetxe-etal-2018-robust, Conneau2017WordTW, Xu2018UnsupervisedCT}. These methods have  been applied to discover cross-language translations of words or produce mappings for bilingual lexicon induction \cite{lample2018unsupervisedmachinetranslationusing,muresanInducingTerminologiesText2013, joulin-etal-2018-loss, artetxe-etal-2016-learning}, which power a range of applications, such as cross-lingual information retrieval systems \cite{litschko2018unsupervisedcrosslingualinformationretrieval}, where a user can input a query in one language, but receive relevant results from documents in other languages, and studying semantic change of terms over time within the same language \cite{kutuzovDiachronicWordEmbeddings2018,kulkarniStatisticallySignificantDetection2015,hamiltonDiachronicWordEmbeddings2018,schlechtwegWindChangeDetecting2019}, or between different sub-communities of the same language, such as across partisan divides \cite{KhudaBukhsh2020WeDS, milbauer-etal-2021-aligning}. 

Because we rarely expect there to be abundant parallel data between pairs of scholarly domains, and do expect domain-specific corpora to be relatively small (if the domains in question are quite granular), we specifically draw on \textit{unsupervised} alignment methods that are robust to small corpora concerns in low-resource settings, such as MUSE \cite{Conneau2017WordTW} and VecMap \cite{artetxe-etal-2018-robust}.

\section{System Design}\label{sec:methods}
Here, we discuss the design and implementation of a prototype cross-domain retrieval system that leverages embedding alignment techniques. The goal of these systems is to enable a user to input a term from a ``home'' domain, and explore sets of terms from a different ``target'' domain that may map to the same or related concepts as the home term. \rev{We overview the functionality of the retrieval system, and then describe how we train and align domain-specific embeddings to facilitate target-community retrieval.} 

\subsection{Cross-Domain Search Engine based on Aligned Domain-Specific Embeddings}

\begin{figure*}[htbp!]
    \begin{center}
    \includegraphics[width=1.5\columnwidth]{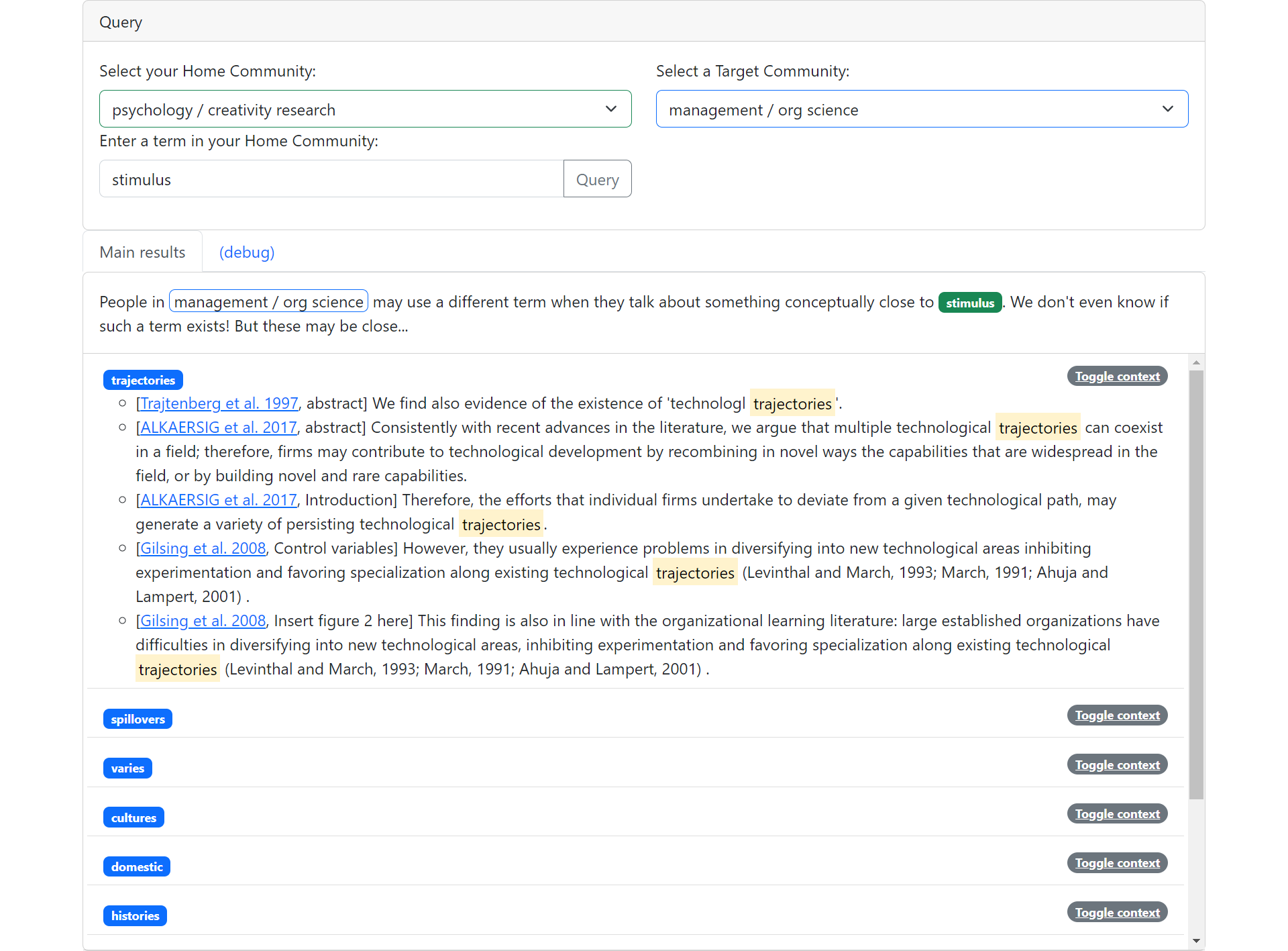}
    \end{center}
    \caption{Term search user interface for exploring potential home $\rightarrow$ target bridges}
    \label{fig:UI}
\end{figure*}

Figure \ref{fig:UI} shows a prototype search engine that could provide this functionality. It enables users to select a ``home'' and ``target'' research domain, to explore how a specific user-defined term from that home domain might align with specialized terms from the target domain. 

The search for related terms from the target community draws on the vocabulary from the target community's domain-specific corpus, and ranks the terms from that vocabulary based on their cosine similarities between their embeddings and the query term in the home-target aligned space. To facilitate sensemaking of results, we retrieve \rev{links anchored with context for the term.} 

\subsection{Training and Aligning Domain-Specific Word Embeddings}


\begin{figure*}[t]
    \centering
    \includegraphics[width=0.5\linewidth]{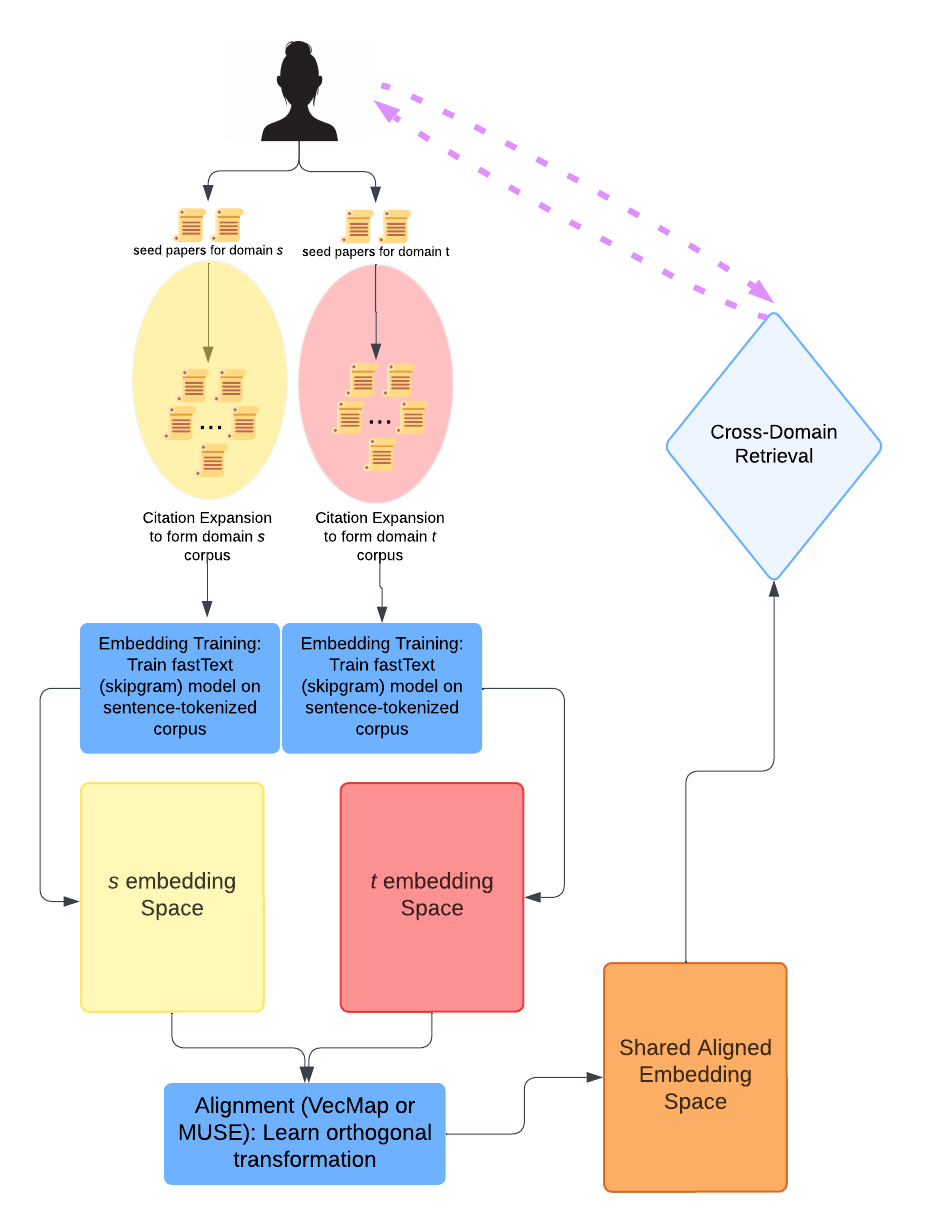}
    \caption{\rev{Overview of the pipeline from researcher-provided seed papers, through corpus development, mono-domain embedding training, and alignment to support retrieval across domains.}}
    \label{fig:pipeline-diagram}
\end{figure*}

To train domain-specific embeddings, we start by constructing separate corpora for different research domains.  Motivated by conceptualizations of research domains as not just disciplines, but potentially ``invisible colleges'' of researchers cohering around common approaches to shared research interests \cite{craneInvisibleCollegesDiffusion1972}, we define domain-specific corpora in terms of citation networks, rather than overall subject headings. Specifically, we seed a research domain by collecting a set of ``seed'' papers for that domain collaboratively with the researchers, and then define a corpus for that research domain by sampling additional papers through citation expansion.

Once we have defined domain-specific corpora, we train domain-specific word embeddings for each domain corpus. For each domain $d_s \in D$, where $D$ refers to all the domains encompassing the set of papers collected for a specific researcher, we train a word embedding model, resulting in an embedding space $E_{d_s}$, which maps words in that domain to vectors in a continuous space. Mathematically, the embedding for domain $d_i$ is a mapping $
E_{d_s}: V_{d_s} \rightarrow \mathbb{R}^k$ from $V_{d_i}$, the vocabulary of domain $d_i$ to a $k$-dimensional vector space, where each $E_{d_i}(w)$ is a vector in $\mathbb{R}^k$ that embeds the usage patterns of the word $w \in V_{d_i}$.

\begin{figure}[htbp!]
    \begin{center}
    \includegraphics[width=0.9\columnwidth]{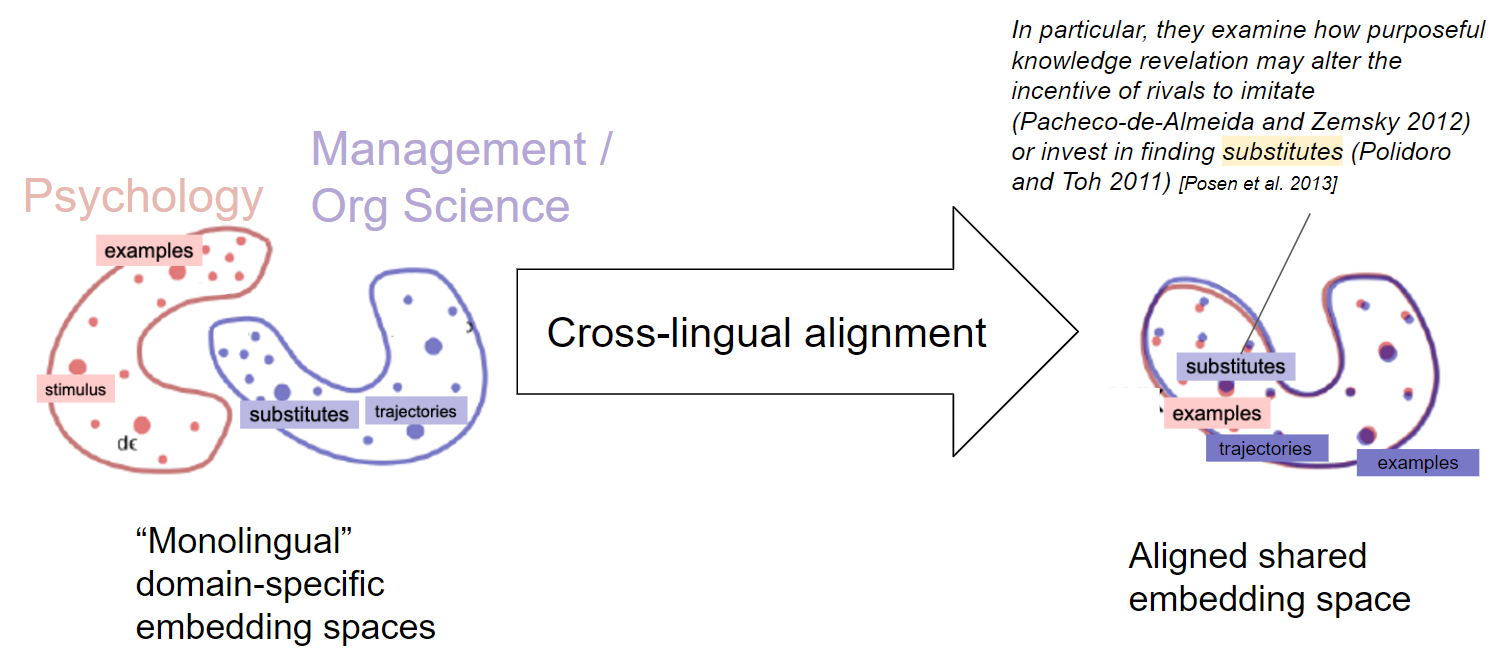}
    \end{center}
    \caption{Visual example using a cross-lingual alignment method to align two research communities (Psychology and Management / Organization Science research on creativity and innovation). The intuition is to find a ``rotation'' that can align two domain-specific embedding spaces such that terms that are semantically similar across the two embedding spaces are close together in the aligned space. Here we show an example from our data, where the MUSE alignment mapped the term ``examples'' (from Psychology) to the conceptually related term ``substitutes'' (from Management) in the aligned space. Visual of rotating spaces adapted from \cite{Conneau2017WordTW}.}
    \label{fig:mapping}
\end{figure}

Our goal is to learn an alignment function $a$ such that $E_{d_t} = a(E_{d_s})$, effectively transforming the embeddings from one domain to align with those of another.  In this work, we consider alignment functions $a$ that are linear transformations of the original embedding space and can be intuitively interpreted as rotations of the embedding space $E_{d_s}$ to better align with the space in ${E_{d_t}}$. Figure \ref{fig:mapping} provides an illustrative example of how similar terms from different domains may become closer in the aligned embedding space. 

In our experiments, we trained fastText\footnote{https://fasttext.cc} embeddings with the skipgram model, a neural network model that learns word embeddings by learning to predict words in a context window around a given input target word.

We then apply computational techniques to learn our alignment function $a$ to \textit{align} pairs of domain-specific embeddings into a shared embedding space. Within this space, we can then explore how terms from one domain-specific corpus --- defined by its embedding in the aligned space --- relates other terms in another domain-specific corpus via metrics such as cosine similarity to the embeddings for those terms in the aligned space. 


We draw our techniques from work on building cross-lingual embeddings (CLE). Specifically, we adapt VecMap \cite{artetxe-etal-2018-robust} and MUSE \cite{lample2018unsupervisedmachinetranslationusing}, both of which learn orthogonal transformation matrices that maintain the internal structure of each monolingual space while aligning them within a shared cross-lingual space. \autoref{fig:mapping} shows that after applying cross-lingual alignment between domain-specific embeddings for ``Psychology'' and ''Management / Organization Science'', the embedding for the term ``examples'' from Psychology could be moved to be closer to the term ``substitutes'' from Management / Org Science in the aligned space. This alignment could inspire new conceptualizations of the process of creating from examples, by considering dynamics of negotation, competition, and strategy (which are central to the concept of strategic substitutes). \rev{We visualize this entire process in \autoref{fig:pipeline-diagram}.} 

\section{Case Study 1}
\label{sec:eval_case_1}

To explore the potential of our alignment approach, we first investigate the novelty and relevance of terms from a target domain retrieved in response to a query term from a ``home'' domain using a MUSE-aligned shared space between the domains. We compare these results to terms 
retrieved by other (monolingual) baseline approaches. We chose an example setting of a researcher studying creativity from the perspective of Psychology, seeking cross-disciplinary insights from Management and Organization Science.

\subsection{Scenario and Domain Corpora Preparation}
We build a corpora of papers with the S2ORC dataset \cite{lo-etal-2020-s2orc}. In this study, a domain is defined by a set of papers relevant to ``creativity'' research in Psychology. To build the corpora, \rev{the fourth} author provided seed papers in Psychology and in Management. A list of all seed papers can be found in the accompanying supplemental materials. We then enlarged the pool of papers for each community through citation expansion: we retrieved papers from a single level of both incoming and outgoing citations from each seed paper. Some statistics of the corpora are provided in Table ~\ref{tab:community-corpus-study1}.

\begin{table*}[h]
    \centering
    \small
    \begin{tabular}{l|cccc}
        Community & \# seed papers & \# papers after expansion & \# sentences & \# tokens \\
        \hline
        psychology / creativity research & 7 & 659 & 77.9k & 47.1k  \\
        management / org science & 8 & 280 & 30.4k & 27.3k\\
    \end{tabular}
    \caption{Corpus statistics for the community corpora used in case study 1}
    \label{tab:community-corpus-study1}
\end{table*}

We constructed a scenario where the psychology researcher is seeking interesting parallels to the concepts of "stimulus" and "examples" from their ``\textbf{home}'' community of Psychology in the \textbf{``target''} community of Management and Organization Science.
The researcher would review a set of terms (in the context of the sentences they were used in) and assess their degree of \textit{relevance} (alignment with their home concept) and \textit{novelty} (degree to which it expanded their thinking about the topic). We chose this initial setting because \rev{the fourth} author of this work has a background in psychology and is studying creativity, and would thus be able to judge the novelty and relevance of retrieved terms.

\subsection{Comparison Points and Evaluation}
To evaluate how our MUSE adaptation might perform compared to existing approaches to search, we constructed an evaluation set consisting of aligned terms and their context from our MUSE approach, as well as 3 baselines that implement some variant of ``monolingual'' embeddings: 
\begin{enumerate}
    \item \texttt{fasttext-target}: using monolingual fastText embeddings trained on the whole corpus (across communities), but retrieving only from the target corpus. This baseline is a direct comparison to the MUSE pipeline, with the only difference being the embeddings
    \item \texttt{fasttext-combined}: using monolingual fastText embeddings trained on the whole corpus, and retrieving from both the home and target corpus, and 
    \item \texttt{SBERT}:  using an off-the-shelf contextual word embedding, retrieving from both the home and target corpus
\end{enumerate}

We expect \texttt{fasttext-target} and \texttt{fasttext-combined} to perform poorly relative to MUSE, but for different reasons: we expect \texttt{fasttext-target} to return results that are contaminated by terms that are similar on the surface, but actually refer to different concepts in the respective scholarly domain; in contrast, we expect  \texttt{fasttext-combined} to retrieve mostly direct matches from the home community. \texttt{SBERT} is a strong baseline, to explore whether existing contextual embedding approaches might be sufficient for teasing out cross-disciplinary nuances in the use of terms. 

Each retrieval pipeline returned 50 (term, context sentence) pairs, where the term occurs in the context sentence. Details of each pipeline follow.


\subsubsection{MUSE, target corpus}
In this pipeline, we leverage the MUSE aligned embedding space.
Given the user's query term, the 10 most similar terms in the target space are obtained.
Then, for each of the 10 expanded terms, 5 context sentences are retrieved from the target community corpus.
Each paper is allowed to contribute up to 2 context sentences; once the maximum is reached, the pipeline would move on to the next paper that has sentences with occurrences of the expanded term. 

\subsubsection{BASELINE 1: fastText, target corpus}
This condition differs from MUSE in that the similar terms are obtained from fastText embeddings.
The context sentence retrieval part is almost the same, except that we may go beyond the top 10 expanded term and use the 11th, 12th, ... similar terms, until we get 50 context sentences.
The context sentence search is over the target community corpus.

\subsubsection{BASELINE 2: fastText, combined corpus}
This is another fastText baseline, which is the same as above except that the context sentence search is over the combined corpus containing papers of all communities.

\subsubsection{BASELINE 3: Off-the-shelf, contextual (SBERT), combined corpus}
This is an off-the-shelf search baseline which does not have the term expansion step.
In contrast, both the query term and all the context sentences are encoded with contextualized embeddings. We retrieve a list of sentence contexts from the target community, then sort it by SBERT cosine similarity with the query term. To fit the (response term, context sentence) format, we assign the word in the context sentence that has the highest universal fastText embedding similarity to the query term in the ``retrieved sentence'' as the ``response term''.
We use the SBERT toolkit\footnote{https://www.sbert.net} and adopt the \texttt{all-mpnet-base-v2}\footnote{https://huggingface.co/sentence-transformers/all-mpnet-base-v2} model.
We consider this as a suitable contextualized embedding model for comparison for this task because SBERT's training data contains the S2ORC dataset \cite{lo-etal-2020-s2orc}.


\subsection{Relevance and Novelty Judgments}
To enable judgments of the results while being naive to their source, we mixed the results from each pipeline into a spreadsheet. The author with expertise in creativity then rated each of the 200 term occurrences (i.e., a term and its context sentence), blind to pipeline source, along two dimensions: 1) \emph{relevance} (0 to 2) and \emph{novelty} (0 to 2).


\subsection{Results}
\label{sec:result}

\begin{figure*}
    \begin{center}
    \includegraphics[width=2\columnwidth]{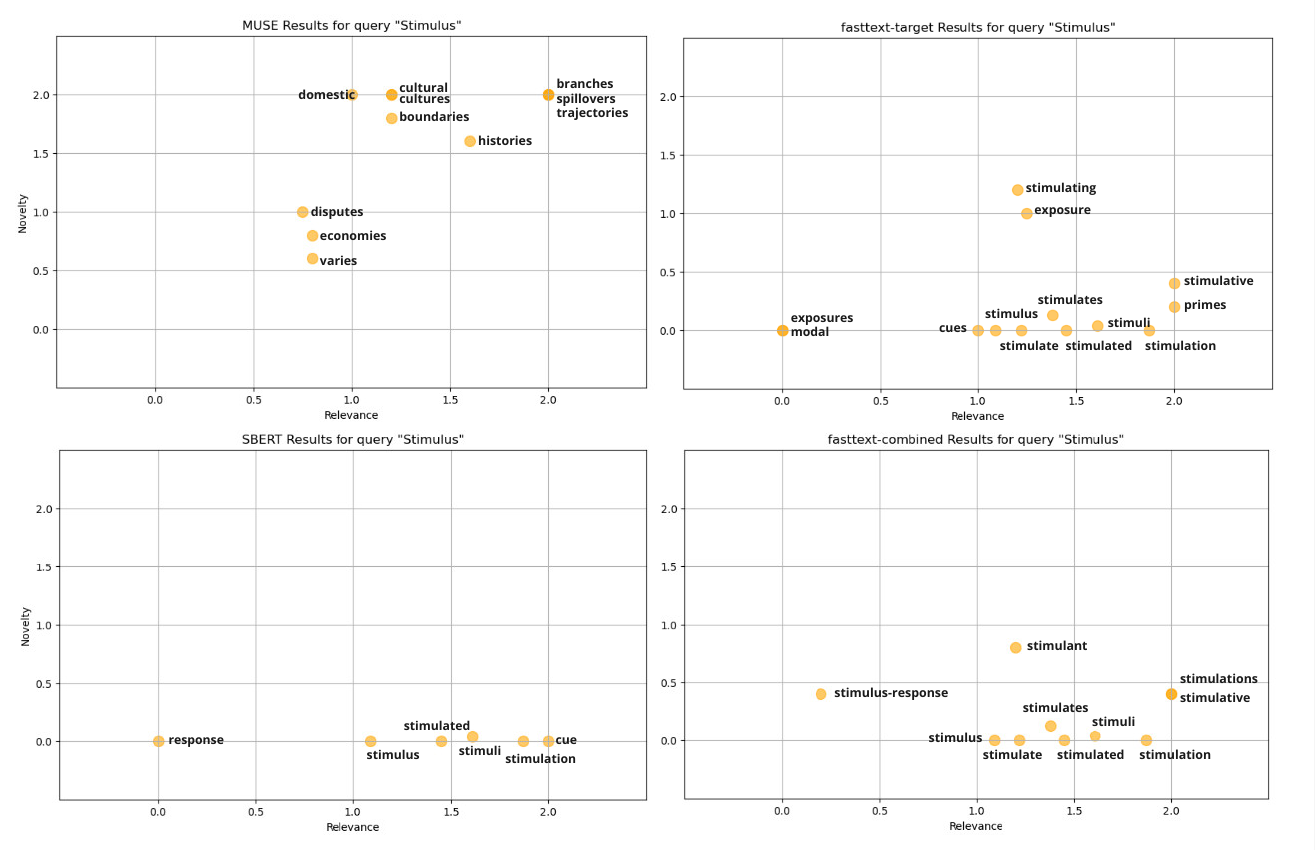}
    \end{center}
    \caption{Results and corresponding relevance and novelty ratings from each pipeline for the query: Psychology[stimulus] $\rightarrow$ Management.  More terms in the MUSE pipeline were rated as having both higher relevance and novelty, compared to other pipelines, which were highly relevant but not as novel. }
    \label{fig:heatmap_1}
\end{figure*}

\begin{figure*}
    \begin{center}
    \includegraphics[width=2\columnwidth]{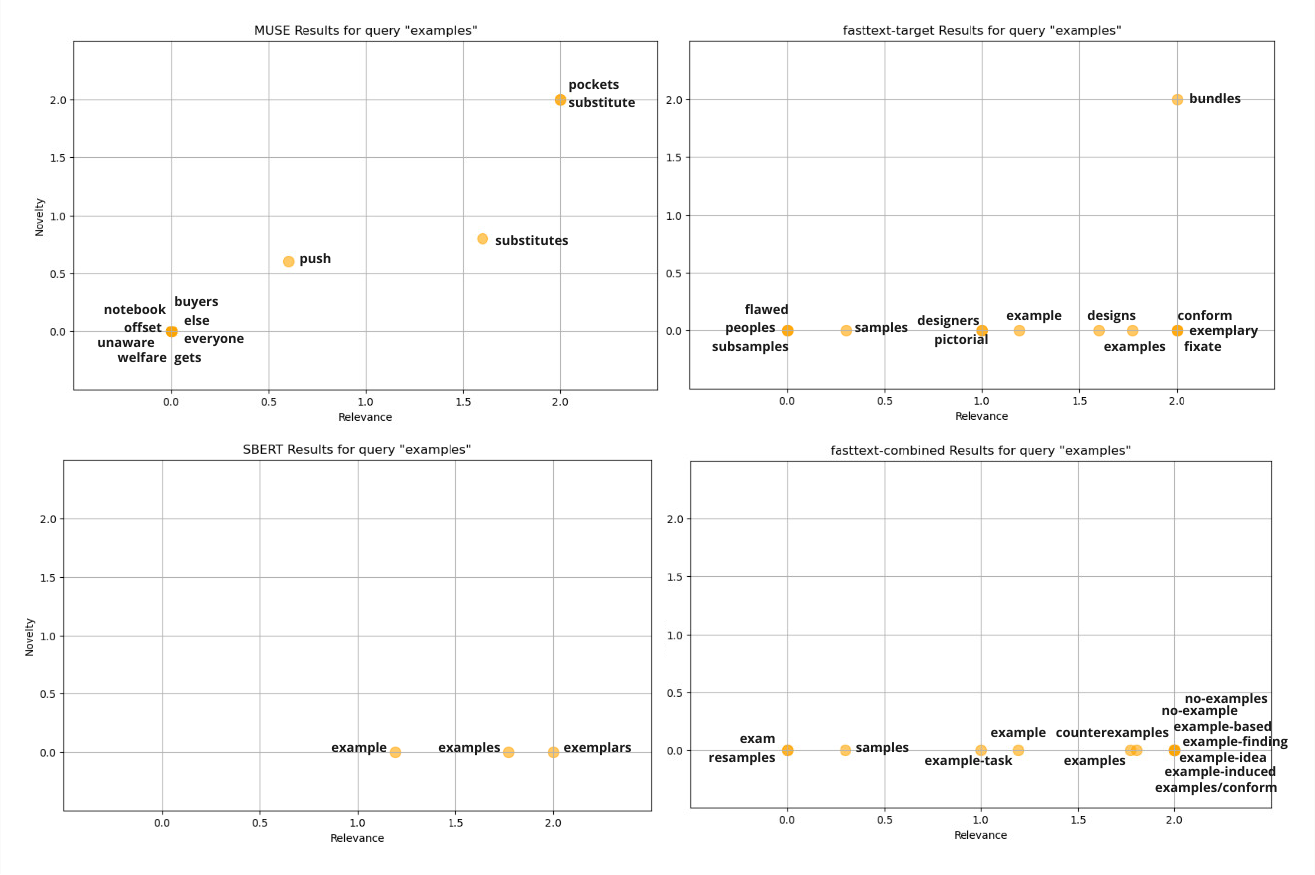}
    \end{center}
    \caption{Results and corresponding relevance and novelty ratings from each retrieval pipeline for the query: Psychology[examples] $\rightarrow$ Management. Across retrieval systems, many of the terms were relevant but not very novel. }
    \label{fig:heatmap_2}
\end{figure*}

Figure~\ref{fig:heatmap_1} and Figure~\ref{fig:heatmap_2} show scatterplots of the relevant and novelty ratings for the query results from each pipeline, with the corresponding terms plotted on the scatterplot for clarity. Each term's relevance or novelty rating is averaged over ratings provided to its specific occurrences in context sentences. Note that the number of terms for SBERT is often less than the other pipelines, since retrieval is focused on \textit{sentences} (and then extracting the most relevant terms), rather than terms directly. These plots show that, on average, the results from MUSE were rated as more novel, while maintaining a high degree of relevance. In contrast, the fastText and SBERT baselines primarily returned relevant results with low novelty. 
This advantage of MUSE over the monolingual baselines was strongly pronounced for the ``stimulus'' query. For the ``examples'' query, MUSE results are higher variance, with high novelty/relevance terms like (evolutionary) ``pockets'', and (strategic) ``substitutes'', but also noisy matches to irrelevant terms like ``notebook'' and ``buyers''. 
But the quantitative pattern of high relevance but low novelty holds for the monolingual baselines.

Qualitatively, the monolingual baselines generally returned results that are very similar, and in many cases, linguistic variants of the original query (e.g., stimuli, stimulus, cue). In some cases, these surface form variations were actually less relevant to the original query term sense: for instance, a substantial number of the results from the monolingual baselines returned context sentences that used the term ``examples'' in a generic sense, not the specific sense of examples used to stimulate creativity. In contrast, the most relevant results from MUSE were frequently specialized terms that are quite different conceptually from the original query term. For example, for the query ``stimulus'', MUSE returned specialized terms like ``spillovers'' (referring to transfer or R\&D knowledge across industries), ``cultures'' (outlining a different dimension of knowledge transfer), and ``trajectories'' (emphasizing the conceptual history of knowledge that can be transferred across domains). For the query ``examples'', key results included specialized terms like (coevolutionary) ``pockets'' (connecting the idea of example-based inspiration to genetic algorithms and evolutionary dynamics), and (strategic) ``substitutes'' (evoking processes of negotiating strategic interests around substitutes for assets and resources in management science).

\section{Case Study 2}
To further explore the potential of our alignment approach, we conducted a second case study with an interdisciplinary researcher in Applied Linguistics, Human-Computer Interaction, and Education, interested in target disciplines of Ethnic-Racial Identity, Professional Identity, and Immigrant Studies. We report the community corpora in \autoref{tab:community-corpus-study2}.

We assess the validity of our tool with a think-aloud study with our prototype cross-domain search engine. In addition to using the prototype to evaluate term alignments, we also use it as a design probe to exploring how our research participant uses the tool to assess related concepts in a target domain. As in Case Study 1, we constructed a scenario where this researcher would see parallels to predefined concepts; here, we use the concepts of ``Education[third space]’’ and ``Applied Linguistics[translanguaging]’’, to explore conceptual alignments with research communities of Ethnic-Racial Identity, Professional Identity, and Education. 

\subsection{Scenario and Domain Corpora Preparation}
We developed the information-seeking scenario and domain corpora for this case study in collaboration with \cspn{}\footnote{Pseudonym to protect participant anonymity}, a PhD student in Human-Computer Interaction with strongly interdisciplinary research interests. The broad question that drove her research was \textit{"How to communicate across different languages for different social and communicative purposes, while being mindful of linguistic dominance dynamics"?} Subthemes and areas of interest for this broad question included how language use relates to dynamics of identity expression, developing intersubjectivity, and transferring knowledge across different language communities. 

In a series of initial conversations with \cspn{}, we identified three research communities that she considered her ``home communities'': 1) HCI, language technologies, and machine translation (hci), 2) applied linguistics (ling), and 3) language education (edu). Each of these home communities contained core specialized concepts that were foundational to her research: for instance, ``AI-mediated communication'' was a central concept of interest in HCI, while concepts like ``translanguaging'' and ``linguistic repertoire'' were core ideas in applied linguistics, and ``funds of knowledge'' and ``third space'' were central ideas from (language) education.  She described being interested in exploring connections of these concepts and her research questions to other research domains, such as ethnic and racial studies (eth), immigrant studies (imm), and professional identity development (pro) in interdisciplinary settings. These research communities thus served as the ``target domains'' for this case study.

We seeded \cspn{}'s home domain corpora by asking her to identify a set of seed papers for the key concepts she identified in each home domain (along with any other background readings she found important). We repeated this process for the target domains of ethnic and racial studies and immigrant studies in a similar manner: seed papers for ethnic and racial studies were drawn from an annotated bibliography for an advanced PhD student whose dissertation was on identity development amongst immigrant teens; and seed papers for immigrant studies were drawn from an annotated bibliography for a research project on immigrant acculturation, led by an Assistant Professor whose research focuses on information ethics and justice, and social inclusion, for immigrants from the African, Afro-Carribean, and Afro-Latinx diaspora in the US. The PhD student and Assistant Professor are departmental colleagues of the fourth author. The fourth author, whose core expertise includes interdisciplinarity, then created a reading list on professional identity development in interdisciplinary settings. They used the LitMaps\footnote{\url{https://www.litmaps.com/}} tool to build out a set of readings in the successively expanded citation networks for a seed set of papers on professional identity in interdisciplinary teams. The seed papers for all domains studied here are provided in the supplementary materials.


From these seed papers, we then developed domain corpora for each target community by gathering forward and backward citations from each community's seed papers using the Semantic Scholar Academic Graph API \footnote{\url{https://www.semanticscholar.org/product/api}}. Table \ref{tab:community-corpus-study2} shows the final corpus statistics for these home and target domain corpora.


\begin{table*}[h]
    \centering
    \small
    \begin{tabular}{l|cccc}
        Community & \# seed papers & \# papers after expansion & \# sentences & \# tokens \\
        \hline
        hci  & 28 & 1.7k & 9.8k & 15.7k \\
        ling & 28 & 8.4k & 43.7k & 53.4k \\
        edu & 46 & 4.5k & 22.7k & 32.1k\\
        eth & 21 & 6.4k & 34.1k & 36.5k\\
        pro & 55 & 4.4k & 24.4k & 30.7k\\
        imm & 43 & 8.2k & 45.85k & 47.8k\\

    \end{tabular}
    \caption{Corpus statistics for the community corpora used in case study 2}
    \label{tab:community-corpus-study2}
\end{table*}

\begin{figure*}
    \begin{center}
    \includegraphics[width=\columnwidth]{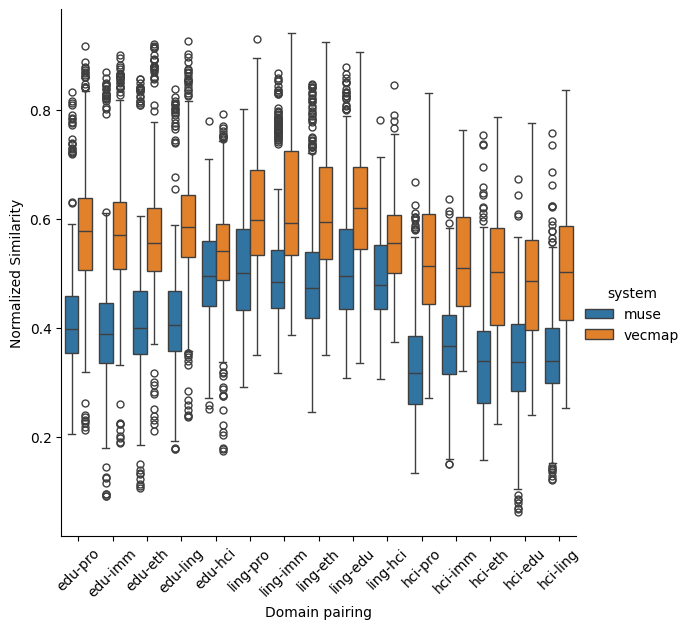}
    \end{center}
    \caption{Pairwise domain evaluation of MUSE and VecMap using normalized cosine similarity on the salient topical terms from the source domain. Salient similarity scores are notably higher with the VecMap alignments, than with the MUSE alignments.}
    \label{fig:quant_eval_case_study_2}
\end{figure*}

We computed normalized modularity score \cite{fujinuma-etal-2019-resource} and present them in \autoref{tab:normalized_modularity_comparison} for several community pairings explored in Case Studies 1 and 2.

\begin{table*}[h]
    \centering
    \begin{tabular}{llcc}
        \hline
        \textbf{Source Domain} & \textbf{Target Domain} & \textbf{VecMap Modularity} $\downarrow$ & \textbf{MUSE Modularity} $\downarrow$ \\
        \hline
        \multirow{5}{*}{hci} 
        & ling & 0.289 & \textbf{0.231} \\
        & edu  & 0.265 & \textbf{0.223} \\
        & imm  & 0.333 & \textbf{0.228} \\
        & eth  & 0.297 & \textbf{0.228} \\
        & pro & 0.229 & \textbf{0.188} \\
        \hline
        \multirow{5}{*}{ling} 
        & hci  & 0.276 & \textbf{0.259} \\
        & edu  & \textbf{0.189} & 0.269 \\
        & imm  & \textbf{0.237} & 0.279 \\
        & eth  & \textbf{0.279} & 0.282 \\
        & pro  & \textbf{0.225} & 0.281 \\
        \hline
        \multirow{5}{*}{edu} 
        & hci  & 0.268 & \textbf{0.228} \\
        & ling & 0.189 & \textbf{0.182} \\
        & imm  & \textbf{0.250} & 0.254 \\
        & eth  & \textbf{0.204} & 0.240 \\
        & pro  & \textbf{0.127} & 0.274 \\
        \hline\hline
        \multirow{1}{*}{psy} 
        & mgmt & 0.469 & \textbf{0.261} \\
        \hline
    \end{tabular}\linebreak \linebreak 
        \caption{Normalized Modularity Comparison between VecMap and MUSE. Lower modularity scores indicate better cross-lingual alignment, and is correlated with higher downstream performance \cite{fujinuma-etal-2019-resource}.}
    \label{tab:normalized_modularity_comparison}
\end{table*}

\subsection{Design Iterations}
We expand on the methodology used in Case Study 1 (\autoref{sec:eval_case_1}) to further explore the application of cross-lingual embeddings for facilitating cross-disciplinary translation. Using MUSE, we started by aligning these embeddings to create the shared vector space where concepts from one domain can be translated into related concepts in a target domain. 

\paragraph{\textbf{Qualitative Sanity Check}}
To understand how well the MUSE system's term mappings aligned concepts between scholarly domains, we conducted a qualitative sanity check. This involved generating a set of term translations using MUSE for several key concepts extracted from both domains. After manually reviewing the generated term mappings we discovered that the mappings did not exhibit the level of coherence we were aiming for, and many terms seemed to lack meaningful counterparts in the target domain. For instance, terms central to one domain’s field of study were often mapped to generic terms and concepts in the target domain, rather than particularly interesting or novel concepts. For example, we observed that  \textit{Education[funds of knowledge]  $\rightarrow$ Ethnic-Racial Identity} would map to terms like ``social psychology'', ``socio-cognitive'', and ``developmental processes''. We also observed broken mappings, where target-mapped terms would not be meaningful, e.g.  \textit{Human-Computer Interaction[AI-mediated communication]  $\rightarrow$ Education[j.]}, where ``j.'' is a response term.


\paragraph{\textbf{Pivot to VecMap}}

We decided to leverage the same monolingual fastText embeddings but experiment with VecMap \cite{artetxe-etal-2018-robust}. VecMap has been shown in cross-lingual contexts to be more robust than MUSE \cite{doval-etal-2020-robustness}, which \rev{can suffer} from instability and stochasticity. Qualitatively, we saw more potentially relevant and interesting conceptual mappings with VecMap, and fewer obviously irrelevant mappings. For example, \textit{ling[repertoire] $\rightarrow$ pro[lifeworld]} and \textit{ling[repertoire] $\rightarrow$ pro[sensemaking]}. Both ``sensemaking'' and ``lifeworld'' can be highly related conceptually to ``repertoire'' because these concepts deal with the way people understand and interpret the world informally. We detail complementary quantitative metrics to assess the strength of the alignment through two reference-free metrics: Normalized Modularity \cite{fujinuma-etal-2019-resource} and Normalized Cosine Similarity, as suggested by \cite{hammerl-etal-2024-understanding}.

\paragraph{\textbf{Normalized Modularity}}
Normalized modularity is a reference-free evaluation metric for assessing cross-lingual word embeddings. It works by measuring how well word vectors from different languages mix in a shared embedding space; the motivation is to detect if intra-lingual similarity (if words are more similar within-language) is greater than cross-lingual similarity (if words are more similar across languages), which is an indication that the embeddings fail to align the semantic meanings of words across languages.

The cross-lingual word embeddings are represented as a graph where words are nodes and edges represent similarity between word vectors using cosine similarity. Modularity then measures the difference between the proportion of edges that connect nodes within the same language, and the random expected proportion of these edges. A high modularity value indicates the presence of distinct language-specific clusters, suggesting poor cross-lingual alignment. A low modularity value indicates weaker separation between languages, suggesting better cross-lingual alignment.

\begin{equation}
Q = (e_s - a_s^2) + (e_t - a_t^2)
\end{equation}\label{equation:pairwise_modularity}
We use \autoref{equation:pairwise_modularity} from \cite{fujinuma-etal-2019-resource} to compute normalized modularity score\rev{s} between pairings of domains, where $e_i$ refers to the fraction of edges that connect words in the embedding space of domain $i$, and $a_i$ refers to the expected number of edges within the embedding space of domain $i$. This score is then normalized:
$Q_{norm} = \frac{Q}{1 - (a_{s}^2 + a_{t}^2)}$

\autoref{tab:normalized_modularity_comparison} presents the results for VecMap and MUSE embeddings across a range of source and target pairings. First, we see that HCI $\rightarrow$ ? tends to have a stronger alignment with MUSE. For Applied Linguistics $\rightarrow$ ?, VecMap generally shows better alignment, with the exception of $\rightarrow$ HCI. Finally, for Education $\rightarrow$ ?, the alignment is roughly equivalent between VecMap and MUSE. All source-target pairings in the queries we test in Case Study 2 indicate lower normalized modularity with VecMap.


\paragraph{\textbf{Salient Cosine Similarity}}
We also computed cosine similarity of domain-salient terms and their translations in a target domain as an intrinsic evaluation metric, while normalizing similarities by the average cosine similarity of the embedding space following suggestions from \cite{hammerl-etal-2024-understanding}. This allows us to assess how tightly terms are mapped to similar terms in the shared embedding space. \rev{Our goal was to ensure that the learned representations were tightly aligned to enable similarity computations between embeddings in the domain pairing.} 

To support this evaluation, we extracted the top-100 salient, topical terms from each domain’s corpus using Latent Dirichlet Allocation (LDA) to use as tests. These terms reflect key concepts within each domain's scholarly discourse. The system then translated these terms into their nearest equivalents in the target domain space using either MUSE or VecMap embeddings.

We compute the cosine similarity for each source and target term pair. 
We normalize this score by the average similarity score of all source-target translation pairs for all vocabulary terms in the source domain. \rev{This quantifies} the overall performance of the system's framework in creating semantically meaningful mappings.

\rev{We} found consistently higher cosine similarity scores for VecMap compared to MUSE (see \autoref{fig:quant_eval_case_study_2}). This complements the other qualitative and quantitative metrics to suggest that the MUSE alignment approach may have struggled to generate reliable mappings compared to VecMap, and provide complementary validation of our choice to run Case Study 2 with alignments from VecMap. 



\paragraph{\textbf{Qualitative Assessments and Query Selection for Think-Aloud Study}}
To get a sense of how effectively the system handles the queries we test across different target domains, we extracted \rev{and qualitatively assessed} 10 response terms for each of the following queries, for all target domains: 

\begin{itemize}
    \item Education: funds of knowledge
    \item Education: third space
    \item Applied Linguistics: translanguaging
    \item Applied Linguistics: repertoire
    \item HCI: AI-mediated communication
\end{itemize}

Two authors of this paper independently rated each response term using discrete labels: \textbf{-1: Not relevant at all, 0: Some relevance but generic or unsure, 1: Very relevant and sensible}.  The Pearson correlation of ratings was 0.53, indicating moderate agreement. Per source domain, target domain, and query, we computed the average of those ratings. We also marked each result as a ``potential hit'' if either rater judged it as potentially very relevant and sensible (i.e., a rating of 1). We then summed the number of ``potential hits'' in each home-query-target domain triple to identify the top four most relevant queries to test in Case Study 2 with the interdisciplinary researcher. We did this to prioritize ``recall'' of potentially useful alignments to validate and discuss with \cspn{} in the case study. 
As Table \ref{tab:case-study-2-query-results} shows, these were:

\begin{itemize}
    \item Education (edu): third space $\rightarrow$ Ethnic-racial Identity (eth) (5 potential hits)
    \item Education: third space $\rightarrow$ Professional Identity (pro) (4 potential hits)
    \item Applied Linguistics (ling): translanguaging $\rightarrow$ Education (5 potential hits)
    \item Applied Linguistics: translanguaging $\rightarrow$ Ethnic-racial Identity (7 potential hits)
\end{itemize}

\begin{table*}[h!]
    \centering
    \begin{tabular}{>{\raggedright\arraybackslash}m{0.75cm}>{\raggedright\arraybackslash}m{2cm}>{\raggedright\arraybackslash}m{1cm}>{\raggedright\arraybackslash}m{1.25cm}>{\raggedright\arraybackslash}m{1.25cm}>{\raggedright\arraybackslash}m{6cm}}
    \hline
    \textbf{Home} & \textbf{Query} & \textbf{Target} & \textbf{Avg. Rating} & \textbf{\# Potential Hits} & \textbf{Potential Hits} \\ \hline

        
  edu & funds of knowledge & eth & -0.45 & 1 & group\_\_identity \\ 
     &  & hci & -0.65 & 0 & \\ 
     &  & imm & -0.20 & 2 & conceptual\_\_framework, tridimensional \\ 
     &  & ling & -0.39 & 1 & multimodality \\ 
     &  & pro & -0.11 & 1 & habitus \\ \hline
     & third space & eth & 0.2 & 5 & social\_\_identity\_\_theories, social\_\_identity\_\_theory, social\_\_identity\_\_aproach, constructionist, identity\_\_negotiation \\ 
     &  & hci & -0.67 & 1 & intersectional \\
     &  & imm & -0.30 & 2 & TDE, information\_\_worlds \\ 
     &  & ling & -0.25 & 3 & multi/plurilingual continua, intertextuality \\
     &  & pro & -0.05 & 4 & CoPs, brokerage, boundary-crossing, brokering \\ 
     \midrule
    hci & AI-mediated communication & edu & -0.75 & 1 & nepantla \\
     &  & eth & -1.00 & 0 & \\ 
     &  & imm & -0.50 & 1 & e-diaspora \\ 
     & & ling & -0.45 & 2 & transnational, transnationalism \\ 
     &  & pro & -0.75 & 0 & \\
     \midrule
    ling & repertoire & edu & -0.35 & 3 & repertoires, repertoire, tacit \\ 
     &  & eth & -0.15 & 4 & heritage\_\_culture, worldview, subculture, heritage \\ 
     &  & hci & -0.95 & 0 & \\ 
     &  & imm & -0.45 & 1 & host\_\_culture \\ 
     &  & pro & -0.22 & 3 & repertoire, lifeworld, sensemaking \\ \hline
     & translanguaging & edu & 0.03 & 5 & translanguaging, teacher-talk, transliteracy, translingual \\ 
     &  & eth & 0.45 & 7 & transcultural, cultural\_\_identity, EICM, multicultural\_\_education, bicultural\_\_identity, ethno-cultural, acculturation\_\_process \\ 
     &  & hci & -0.17 & 3 & mixed-initiative, cooperative, co-creative \\ 
     &  & imm & -0.45 & 0 & \\ 
     &  & pro & 0.00 & 3 & situated\_\_learning, CID, situated\_\_learning\_\_theory \\ \hline
    \end{tabular}
    \caption{Qualitative assessment results terms returned for all home-domain[query]-target-domain pairs tested for Case Study 2, assessed by two authors of this paper. To prioritize recall for validation in the think-aloud, potential hits are defined as results where at least one rater gave a 1 (very relevant and sensible).}
    \label{tab:case-study-2-query-results}
\end{table*}

We selected domain-query pairs with a higher proportion of potential hits because terms that were irrelevant or too broad were quite straightforward to recognize, but we wanted expert validation of the terms that we saw as potentially meaningful analogs in the target domains, and could seed meaningful discussion on cross-domain exploration.

\subsection{Study Design}\label{subsec:experimental_design_study_2}
\paragraph{\textbf{System Comparison}}
In our second case study, we introduced a baseline comparison using GPT-4o-mini alongside the VecMap system. Our goal in comparing to GPT is not to assess if our system could outperform an LLM, but rather to use it as a point of comparison for an alternative approach that relies on much larger scale data and compute, and for which we already have some evidence of multilingual capabilities \cite{Kew2023TurningEL, armengol-estape-etal-2022-multilingual}. 

We conducted a blind comparison between the two systems using a repeated measures experimental design with our interview participant, aiming to assess their identification of sensible, and sensible+new terms. For each query, our participant viewed the results of each system in a counterbalanced order to mitigate ordering effects. This experiment design allows us to directly compare the performance of the two systems in generating meaningful and relevant terms across different interdisciplinary domains. 

The results for each query were shown in our prototype cross-domain search engine interface \autoref{fig:UI}: both GPT and VecMap results could be displayed in an identical fashion, showing a ranked list of terms and a set of context sentences for each term, along with URLs to the research papers that contain those sentences, for more context. For each set of results for a query, we asked the participant to identify and discuss any terms that were sensible and sensible+new. After this stage, we allowed for free play of the VecMap system to spur needfinding discussion on the cross-domain information seeking task.  The entire case study was conducted in-person, and lasted approximately 60 minutes. During this time, the participant's screen and discussion when navigating the system was recorded using Zoom. The recordings were later transcribed using MacWhisper \footnote{https://goodsnooze.gumroad.com/l/macwhisper} for analysis.

We discuss in more detail how the retrieval pipeline for the two systems are implemented below, and then jump into the results.

\paragraph{\textbf{VecMap Alignment System}}
With this system, we leverage the VecMap-aligned embedding space. Given the user's query term, the K=10 most similar terms in the target space is obtained. Then, for each of the 10 response terms, up to 5 context sentences are retrieved from the target domain corpus of abstracts. Each paper is constrained to at most 2 context sentences.

\paragraph{\textbf{GPT-4 Baseline}}




\rev{We initially prompted GPT4o-mini to generate related concepts from the target community, but it sometimes produced out-of-vocabulary terms or combined words in unintended ways. Since our search engine relies on adherence to the target community's terminology, we shifted from open-ended generation to constrained selection. Prompt:}

\begin{quote}
You are a researcher in the field of \{SOURCE\_COMMUNITY\}, researching the concept of \{HOME\_CONCEPT\}.
Find 10 similar concepts to this in the field of \{TARGET\_COMMUNITY\}, separated by string comma.
You are ONLY allowed to select from the following terms: \{candidate\_terms\}.
Do not combine different words in this vocabulary, you are ONLY allowed to select from them.
\end{quote}
We note however that this does require larger context windows, to fit the entire vocabulary into the prompt.





\subsection{Results: Think-Aloud}
We discuss the results from the case study as described in \autoref{subsec:experimental_design_study_2}. In \cref{fig:term_comparison_third_space_1,fig:term_comparison_third_space_2,fig:term_comparison_translanguaging_3,fig:term_comparison_translanguaging_4}, we present as a reference the list of terms that the participant indicated were sensible and sensible + new, broken down by GPT and VecMap-based systems. We discuss further insights from each query in \cref{subsec:query_insights_1,subsec:query_insights_2,subsec:query_insights_3,subsec:query_insights_4}.

\begin{table*}[htbp]
    \footnotesize

\begin{center}
    \textbf{Education[third space] $\rightarrow$ Ethnic-racial Identity} \vspace{0.5cm} 
\begin{tabular}{p{1cm}|p{7.5cm}|p{6cm}|} 
    \rowcolor{yellow!20}
    & VecMap & GPT \\\hline
     Sensible + New & 
   \begin{itemize}[leftmargin=*] \item The article concludes that national character is a sociocultural and psychological indicator that influences \hl{identity construction} in political autobiographies.\end{itemize} 
   & \begin{itemize}[leftmargin=*] \item  Parental \hl{cultural socialization} and general parenting quality are important predictors of ethnic identity (EI) development in adolescents. \end{itemize} \\ \hline
   Sensible & \begin{itemize}[leftmargin=*] \item 
   It offers a nuanced understanding of the author's career development journey to an authentic work \hl{identity}. This analytic autoethnography, situated in multicultural, democratic South Africa, describes how historic moments in the country's political evolution influenced the author personally $\cdots$ I am a Black South African; I am a gay professional and so, who am I at work?
   \item This research reexamines common conceptualizations of ethnic identity as a relatively stable, dispositional-like construct—in contrast to \hl{constructionist} theories that emphasize the contextual, situational, and transitory nature of ethnic identity—and integrates the two frameworks.
   \item The current study explored ways schools facilitate supportive or marginalizing experiences for first generation Arab heritage youth in the United States and investigated how these experiences impact acculturative experiences and \hl{identity negotiation} for these students.
   \item The \hl{social identity} framework suggests that exposure to high-status ingroup or low-status outgroup portrayals enhances self-esteem through positive ingroup distinctiveness.
   \end{itemize}  &
   \begin{itemize}[leftmargin=*]    \item The \hl{social identity} framework suggests that exposure to high-status ingroup or low-status outgroup portrayals enhances self-esteem through positive ingroup distinctiveness.
   \item When non-native English speakers enter the classroom, immediate differences in language, cultural values, and peer interactions impact \hl{identity development} and negatively influence self-perceptions. \end{itemize} \\ \hline
\end{tabular}
    \caption{Terms selected by \cspn{} as sensible, and sensible+new for Education[third space] $\rightarrow$ Ethnic-racial Identity. Terms supplied by the retrieval system (either GPT or VecMap) are provided in-context, using just one representative line for demonstration purposes.}\label{fig:term_comparison_third_space_1}

\end{center}
\end{table*}

\begin{table*}[htbp]
    \footnotesize
\vspace{0.5cm}
    \textbf{Education[third space] $\rightarrow$ Professional Identity} \vspace{0.5cm} 

\begin{center}

\begin{tabular}{p{1cm}|p{9cm}|p{3cm}|}
        \rowcolor{yellow!20}

  &  VecMap & GPT \\\hline
       Sensible + New & \begin{itemize}[leftmargin=*] \item Rooted in a sociological model illustrating interacting levels of power at macro-, meso-, and microlevels, we argue that authentic \hl{research-practice} partnerships and clinician-researcher collaborations can mitigate this power differential. \end{itemize}  & ...
     \\ \hline
    Sensible & \begin{itemize}[leftmargin=*] \item Drawing on social learning theory, and in particular on communities of practice (\hl{CoPs}), the results of the study unpack how the digital nature of online peer discussion groups enables three interrelated learning mechanisms (sharing, symphonizing and shaping). 
    \item This study explores how doctors perform this bridging work, conceptualized as cultural \hl{brokerage}, on the job.
    \item Using a \hl{boundary-crossing} lens, the article analyses the workplace arrangement, in which students’ learning across professional boundaries and their negotiations around a boundary object were prerequisites to coordinate their interprofessional knowledge and manage emerging challenges while being in charge of care on the ward.
    \item This article argues that time plays a crucial role of \hl{brokering} disparate knowledge spaces in Indigenous discourses of postgenomics, with both Indigenous cosmological frames and DOHaD/epigenetics centring a circular temporal model.
    \end{itemize} & ... \\ \hline

\end{tabular}
\end{center}

    \caption{Terms selected by \cspn{} as sensible, and sensible+new for Education[third space] $\rightarrow$ Professional Identity. Terms supplied by the retrieval system (either GPT or VecMap) are provided in-context, using just one representative line for demonstration purposes.}\label{fig:term_comparison_third_space_2}

\end{table*}

\vspace{0.5cm} 

\begin{table*}[htbp]
    \footnotesize
\begin{center}
\textbf{Applied Linguistics[translanguaging] $\rightarrow$ Education} \\[0.25cm]

\begin{tabular}{p{1cm}|p{6cm}|p{7.5cm}|}
        \rowcolor{yellow!20}
    & VecMap & GPT \\\hline
        Sensible + New &     
    \begin{itemize}[leftmargin=*] \item Seven bilingual teacher candidates implemented a strategic \hl{transliteracy} approach to writing instruction, observing student assets and teaching cross-linguistically. \item Students can develop a \hl{translingual} awareness that perceives languages as plural and fluid, and this awareness can help students succeed outside of monolingual norms. \end{itemize}  & \begin{itemize}[leftmargin=*] \item Seven bilingual teacher candidates implemented a strategic \hl{Transliteracy} approach to writing instruction, observing student assets and teaching cross-linguistically. \item The present paper aims at finding out what modifications in translation of visitors’ rules may be necessary if the target readership is to be considered, and at explicating the translation process through applying a \hl{transdisciplinary} perspective of ELF studies, linguistic landscape (LL) studies, cross-field studies on conceptualization, translanguaging and translation studies. \item For this purpose, the text addresses, first, the theoretical developments of the decolonial turn; secondly, it discusses the notion of critical \hl{interculturality}; and finally, it presents a reflection on the links between critical interculturality, decolonial pedagogy, and a vision of teacher education as a dynamics of personal development. \end{itemize}

    \\ \hline
    
    Sensible & \begin{itemize}[leftmargin=*] \item The mothers and their daughters drew on their everyday ways of being (e.g., playfulness, \hl{translanguaging}) in their mathematical interactions. \item We argue that our cases show how it is possible to bridge the gap between heteroglossic conceptions of language and \hl{languaging} in applied linguistics, and what is conceived as legitimate language practices in the classroom. \end{itemize} & 
\begin{itemize}[leftmargin=*] \item The mothers and their daughters drew on their everyday ways of being (e.g., playfulness, \hl{translanguaging}) in their mathematical interactions. \end{itemize} 
    
     \\ \hline

    \end{tabular} \end{center}
    
    \caption{Terms selected by \cspn{} as sensible, and sensible+new for Applied Linguistics[translanguaging] $\rightarrow$ Education. Terms supplied by the retrieval system (either GPT or VecMap) are provided in-context, using just one representative line for demonstration purposes.}\label{fig:term_comparison_translanguaging_3}
    \end{table*}


\begin{table*}[htbp]
    \footnotesize
\begin{center}
\textbf{Applied Linguistics[translanguaging] $\rightarrow$ Ethnic-racial Identity} \\[0.25cm]

\begin{tabular}{p{1cm}|p{9cm}|p{3cm}|}
    \rowcolor{yellow!20}

   & VecMap & GPT \\\hline
       Sensible + New & \begin{itemize}[leftmargin=*] \item This study assessed the ability of a set of practice-based measures to identify indicators of positive \hl{transcultural} parenting. \item Becoming multicultural through early immersive culture mixing (\hl{EICM})—i.e., growing up with a mix of cultures that coexist and interact to form an emergent hybrid culture within one’s home—is a rapidly rising phenomenon in many parts of the world. \end{itemize} & ...  \\ \hline
       
    Sensible  & \begin{itemize}[leftmargin=*] \item The \hl{acculturation} aspect examined unique manifestations of conflict related to the \hl{acculturation} process of immigrant families. \end{itemize} & ... \\ \hline

\end{tabular}
\end{center}
\caption{Terms selected by Jane Foster as sensible, and sensible+new for Applied Linguistics[translanguaging] $\rightarrow$ Ethnic-racial Identity. Terms supplied by the retrieval system (either GPT or VecMap) are provided in-context, using just one representative line for demonstration purposes.}
\label{fig:term_comparison_translanguaging_4} 
\end{table*}

\subsubsection{Query Insights: Education[third space]$\rightarrow$ Ethnic-racial Identity} \label{subsec:query_insights_1} (\autoref{fig:term_comparison_third_space_1})

 \cspn{} began by mapping the concept of ``third space'' from Education to Ethnic-racial Identity. 
She emphasized that VecMap’s results were more focused and contextually relevant. Terms like ``social identity'', ``identity development'', and ``cultural socialization'' were sensible and aligned well with her understanding of ``third spaces'' as informal settings of daily life interactions. In contrast, the terms returned by the GPT system, while relevant, focused more on ``racial'' and ``ethnic'' aspects of identity formation. While intrigued by the term \hl{hybrid}, she felt that its association with identity was too distinct from the concept of third spaces:
\begin{displayquote}\textit{
I think this one surprised me a little. I never considered \hl{hybrid} to modify identities. $\cdots$ if it's used in a general sense, let's say social interaction, it may potentially map to ``third space.'' Because it can be a \hl{hybrid} mode of interaction, like it's not entirely in-person or it's not entirely online... Or it's a \hl{hybrid} between family space and other, non-family space. Here, what I'm seeing is a \hl{hybrid} type of identity. So in the identity sense, I wouldn't use this term... But for modality of interaction, yeah.}
\end{displayquote}





\subsubsection{Query Insights: Education[third space] $\rightarrow$ Professional Identity}\label{subsec:query_insights_2} 
(\autoref{fig:term_comparison_third_space_2})

In the second query, \cspn{} again explored
the concept of ``third space'' in education, but this time in Professional Identity. She noted that to her, she would not map ``third spaces'' to professional identity in general, because ``third spaces'' are life-relevant, situated in informal, everyday environments, while professional identity is shaped in more structured, work-related contexts. 

Nevertheless, VecMap yielded terms such as  \hl{research-practice}, \hl{brokering}, \hl{brokerage}, and \hl{boundary-crossing}, which resonated with her idea of third spaces as transitional zones:

\begin{displayquote}\textit{
Ah, I see ... This is sort of a third space, it's between research and practice, somewhere in between ... I have not seen \hl{research-practice} before ... If it's \hl{brokering} or \hl{boundary-crossing}, then it's between two different workplaces...}
\end{displayquote} 


GPT, however, returned terms closely tied to professional settings that were less relevant because they did not capture the nuances of the life-relevant aspects central to ``third spaces'', contrasting them with the terms she found from the VecMap system:
\begin{displayquote}\textit{
 I think because these terms focus more on the professional... I'd consider that most of them in the ``workplace,'' rather than the ``third space.'' Unless it touches upon ``research-practices'' where it maps to another space or in-between... Or if it's ``brokering'' or ``boundary-crossing,'' then it's between two different... [work]spaces. I think if it's professional identity-related, I wouldn't consider these sensible.}
\end{displayquote}

\begin{displayquote}\textit{
I think the only thing is identity formation, so I want to see more in how it is used, and whether it really has an analogical relationship with third space.}
\end{displayquote}


\subsubsection{Query Insights: Applied Linguistics[translanguaging] $\rightarrow$ Education}\label{subsec:query_insights_3} 
(\autoref{fig:term_comparison_translanguaging_3})

For the third query, \cspn{} explored how ``translanguaging'' mapped to terms in  Education. Both systems contributed useful terms, but with different perspectives: VecMap focused more on learning-related concepts, while GPT focused on multilingualism.


While neither system was fully comprehensive, both offered inspiring concepts for the participant. \cspn{} highlighted that translanguaging relates to integrating separate linguistic systems within a person's repertoire, which maps closely to terms like ``transliteracy'', ``transdisciplinary'', and  ``interculturiality''. She noted that related terms like ``code-switching'', although relevant, refers to a distinct linguistic concept from translanguaging.

Same-term responses prompted \cspn{} to consider it in different contexts:
\begin{displayquote}\textit{
It returned translanguaging... but it might be a different context!... Oh it's the same one. I think it's sensible because it's the same one so I'll still put it down... It didn't change my thoughts on translanguaging.} 

\end{displayquote}

\cspn{} was particularly interested in \hl{transliteracy}, \hl{transdisciplinary}, and \hl{interculturiality}, which she felt captured the ``translanguaging'' concept, contrasting those terms with \textit{distinct} terms such as \hl{code-switching}:
\begin{displayquote}\textit{
I see a very strong analogical connection, because 
everything is integrated ... people are just drawing from this one repertoire. When I am considering \hl{transliteracy, transdisciplinary, interculturiality},
it's really: how do we pool everything together and draw connections between them? I didn't include \hl{bilingual} / \hl{pluralinguism}. These are relevant terms but are different concepts from ``translanguaging''... People would think it equates to \hl{code-switching}, but I don't see it that way. These terms in general emphasize: you speak two languages, and how do you switch back and forth? Instead of seeing everything as one unit.}
\end{displayquote}




\cspn{} also assessed the nuances of the ``translanguaging'' term, noticing that some of the response terms were too broad to capture the original concept:
\begin{displayquote}\textit{
I think they're losing the core idea of ``translanguaging''. ``Translanguaging'' COULD be a strategy used to \hl{scaffold} learning, but ``scaffold'' would map to ``translanguaging'' as a ``pedagogy'' instead of ``translanguaging'' itself.}
\end{displayquote} 

\begin{displayquote} \textit{
I could see ``translanguaging'' as a practice of \hl{teacher-talk}, but it doesn't map there. It's hierarchical $\cdots$ it's \hl{teacher-talk} then underneath it's ``translanguaging.''}
\end{displayquote}






\cspn{} found both systems useful but in different ways. The VecMap system returned terms that were more focused on learning, while the GPT-based system selected terms that were more aligned with language and multi-linguism. While neither system was fully comprehensive, both offered inspiring concepts for the participant. 

\subsubsection{Query Insights: Applied Linguistics[translanguaging] $\rightarrow$ Ethnic-racial Identity}\label{subsec:query_insights_4} 
(\autoref{fig:term_comparison_translanguaging_4})

In the last query, \cspn{} explored how translanguaging maps to Ethnic-racial Identity. While VecMap did not surface many interesting terms, \hl{EICM} \rev{(early immersive culture mixing)} in particular captured her interest. She perceived the term as suggesting an environment where cultural elements are deeply intertwined, making it difficult to separate distinct cultural identities. Within the sentence context of \hl{EICM}, ``hybrid''  resurfaced, connecting back to her first query, which prompted her to reflect further on how hybrid cultures manifest:
\begin{displayquote}\textit{
 I feel like I need to do more reading on ``hybrid''. I'm intrigued by this term in a neutral way. I can't see how it would be related, but because I have seen this... Hybrid of multiple identities; that is a clear delineation. But this is an emergent hybrid. It's hard to see one person's cultural root and then separate it; for example, that part belongs to US culture, that belongs to Chinese culture; so they are using a different.. way [of looking at it].}
\end{displayquote}

As for GPT, thematically, these terms were considered too broad:
\begin{displayquote}\textit{
This focuses a lot on identity in general. Because it's general, it has lost the nuances of ``translanguaging''...}
\end{displayquote}




When prompted to compare the two systems, \cspn{} consistently favored the VecMap-based system, noting that the system was able to deliver more research-focused terms.

\subsubsection{Free Play}

For the free play portion of the system, \cspn{} explored \textbf{Linguistics[translanguaging] $\rightarrow$ HCI} using just the VecMap system. Her goal was to understand how the HCI community discusses the concept of translanguaging and how it might be applied in design contexts.

Some of the response terms did not capture the fully-formed use cases the participant had in mind. For example, \hl{speculative} appeared instead of \hl{speculative design}. Moreover, the system did not offer direct, concrete applications for design. While these terms do not directly map to the core concept of ``translanguaging'', they still could be useful to describe how one might approach studying it in HCI, serving as a guide to consider different ways to approach ``translanguaging'' methodologically.
\begin{displayquote}\textit{
I feel like it gives me a methodological approach, rather than a design application. I don't necessarily think [\hl{speculative}] maps to the concept [of ``translanguaging''], but it's useful to consider my approach... \hl{co-creative} could also be a participatory design method. Reading this list, it seems that HCI would consider ``translanguaging'' as an open question. That there is not a specific or concrete claim yet, but it maps to a series of features of this term that guide the methodological choices to study this term in HCI.}
\end{displayquote}

\subsubsection{Design Feedback}
Discussion with the participant after reviewing queries yielded insights about how a system that enables cross-domain alignment of concepts might support specific translation work scenarios. 

One use case the participant mentioned for exploring unfamiliar domains was to investigate how theories they are familiar with are specifically applied in different target contexts. 

\begin{displayquote}\textit{
If I have good familiarity on a term, I want to see what are the other claims surrounding this topic just so I would understand this term more directly. My perspective is this: this is how people in Linguistics lay claims on this term, and this is the social science theory to understand HCI problems.}
\end{displayquote}

\begin{displayquote}\textit{
I want to know how ``translanguaging'' could be implemented. I know well about the theory, but outside of the classroom, how do we do that? With these terms like \hl{transliteracy}, I want to see how these sort of actions of pooling resources together would work in other contexts, and to see if there are any strategies or behaviors or interventions I could sort of mimic.}
\end{displayquote}

In this scenario, the scholar would query the term ``translanguaging'' from Applied Linguistics $\rightarrow$ Education to get a sense of how the concept of translanguaging is studied in education contexts. This use case can be handled in current search systems in the case that the specific term is used in the exact same way in the target domain. 

A second use case focuses on identifying concepts that are \textit{similar} to a specific concept of interest. As an example, the participant described the need to 
find theoretical backings cited in their home community.
\begin{displayquote}\textit{
So now I would want to see a connection of how I could link the social science theory back to HCI, and I want to get a sense of… ``have similar things been done in HCI?’’ Or explore how the HCI community thinks of this concept, for example, translanguaging [from Linguistics].}
\end{displayquote}

This is akin to the idea of searching for similar theories that inform work done in applied domains, and writing related works to get a sense of what similar things were done in the past. Importantly, this use case of finding alignments between similar concepts may not be not directly supported by techniques that gloss over jargon or domain differences in language use. Instead, methods like cross-domain alignment, as we explored here, might be more useful.

\section{Discussion}
\label{sec:discussion}

In this paper, we explored how we might adapt techniques for cross-domain alignment of word embeddings to support translation work in cross-domain scholarly information seeking.
We then tested this concept across two case studies with interdisciplinary scholars' real cross-domain information seeking scenarios. In the first case study, training and aligning domain-specific word embeddings with MUSE enabled retrieval of more relevant \textbf{and} novel response alignments between terms in home and target research domains, compared to either the contextual SBERT approach and the combined-corpora modeling approaches (more relevant, less novel). 
In the second case study, 
training and aligning domain-specific word embeddings with VecMap enabled retrieval of equally or more relevant and novel alignments between terms in home and target research domains, compared to responses from GPT. Because LLMs have been trained over diverse, large-scale sources of data, we analogize this setting to the combined-corpora modeling approaches we tested in Case Study 1. However, the results of Case Study 2 point to \rev{off-the-shelf} LLMs being unable to produce terms suitable for this cross-domain exploration task for scholars. We note the effectiveness of our prototype approach, despite the small size of the underlying corpora. 

\rev{Together, these results suggest that modeling communities separately and then aligning them into a shared embedding space can preserve specialized knowledge and produce opportunities for interesting concept explorations.} Overall, we view the quantitative and qualitative example results as a promising proof of concept that multilingual NLP approaches might be fruitfully applied to the scenario of cross-domain information seeking, by framing scholarly domains as distinct language-using communities. 

\subsection{Limitations and Future Work}

Our case studies provide high ecological validity and qualitative depth, due to their grounding in real cross-domain information seeking scenarios and direct involvement of interdisciplinary scholars; but more precise quantitative estimates of efficacy across additional settings are needed to establish generalizability and robustness.

Another important next step will be to systematically investigate the conditions under which different alignment-based approaches are likely to be fruitful. Our investigation here suggested that VecMap was a more effective alignment technique for the Case Study 2 corpora compared to MUSE, but we have little direct insight into precisely why this is the case. Further, having seen the stronger results for VecMap in Case Study 2, we attempted to replicate and improve on the results of Case Study 1 with VecMap replacing MUSE as the alignment technique: unfortunately, the resulting alignment failed to produce similarly relevant and novel terms as MUSE. Investigating the reasons for these performance differences is an important next step. This could begin by investigating specific assumptions made by alignment techniques about the size, distribution, and overlap between domain corpora, but also investigate how these interact with specific alignment technique hyperparameters, such as the choice of vocabulary size and dimensionality of embeddings to keep.

These investigations will be greatly enhanced by the availability of reliable intrinsic, automated evaluation metrics to correlate with the extrinsic cross-domain translation task we are interested in. In our investigations, we adapted evaluation metrics that show moderate correlation with cross-lingual tasks, but there is no prior evidence on their effectiveness in this specialized task. Thus, our choice to pivot to VecMap for Case Study 2 was primarily driven by our qualitative manual inspection of query results: the cosine similarity of salient terms and normalized modularity metrics \cite{fujinuma-etal-2019-resource} were interpreted overall as a complementary validation of our choice, rather than a fine-grained evaluation of the quality of specific domain alignments. 
We also observed some correlations between these metrics and the qualitative assessment ratings, in theoretically predicted directions: normalized modularity was negatively correlated with average qualitative rating scores, $r=-.37$, and salient cosine similarity was positively correlated, $r=.57$. These observations suggest these metrics may be useful for automatically evaluating alignment quality for this task. However, we are hesitant to interpret these as strong quantitative results, given the small N (15 total domain pairs, 1-2 queries per pairing), and the lack of gold standard ratings for the query results (since the domain expert only directly validated four sets of query results).
Thus, there is still a need to directly validate alignment metrics for this specific application of aligning different scholarly domains.

\rev{Finally, while both our case studies benefit from the involvement of PhD-level scholars with the relevant interdisciplinary expertise, the selection process introduces certain limitations. We chose these two participants based on their professional relevance to different case study domains, limiting our ability to generalize the findings from the qualitative evaluation of the cross-disciplinary term alignments to the diversity of scholars in these fields.}  


\subsection{Additional Multilingual NLP Approaches to Explore for Cross-Domain Scholarly Translation Work}

In this paper, we tested the concept of alignment-based approaches with the simplest instantiation: linear alignment of monolingual embeddings, and computing similarities between word embeddings in the shared space. Despite the promising examples of useful alignments, the quantitative trends (e.g., 50-60\% potential hits at 10 results on the upper bound, but a number of cases with little to no strong potential results) indicate the difficulty of the task. Thus, we are excited to explore additional extensions and multilingual NLP approaches that could support cross-domain scholarly translation work. 

One possible direction is to move away from a strictly unsupervised setting, and adapt additional techniques from multilingual alignment to improve alignment quality, both at the alignment stage and the retrieval stage. First, we might be able to enhance the quality of our cross-domain scholarly alignment by injecting supervision into learning better representations. For example, labeled pairs of specialized terms from different domains can help bootstrap better alignments. Additionally, semi-supervised techniques such as self-training could help expand coverage of cross-domain mappings, allowing us to benefit from the flexibility of the unsupervised approach and the effectiveness of more supervised techniques. From a retrieval perspective, we could integrate techniques from informational retrieval, such as learning-to-rank, or ensembling our distributional semantics-based approach with traditional retrieval methods (e.g. BM25 and TF-IDF). Another direction might be to use provided or learned \cite{forer2024inferring} hierarchies from ontologies or controlled vocabularies, to focus retrieval on more specific (vs. broader) domain-specific terms.


Finally, it may be fruitful to explore multilingual approaches powered by LLMs. In Case Study 2, we tested a prototype translation-like prompting approach with GPT-4o to select potentially relevant and novel terms within the constraints of our setup. While initial results seemed promising, the think-aloud results revealed that terms were still too imprecise and broad. Natural extensions could be to refine the prompt design to explore how to select novel and sensible terms; some prompting-based methods could range from: few-shot prompting, chain-of-thought, to prompting pipelines on top of embedding-based retrieval.








\subsection{Implications for Intelligent Interactive Systems for Supporting Cross-Domain Scholarly Information Seeking}

From a systems perspective, we are keen to explore new interface designs for reading and searching that draw on multilingual techniques to support cross-domain information seeking. 
One example might be a ``translate'' button in a reading interface: after providing information about their ``home'' domain, scholars could select text portions that are unfamiliar to them, right click to show the menu, and then click ``translate'' to see term mapping to their home domain. This could be implemented in a user interface similar to the \texttt{ScholarPHI} system for augmenting scientific papers with position-sensitive \textit{definitions} of mathematical terms and symbols \cite{headAugmentingScientificPapers2021}. Another approach might be to augment existing search systems with a list of cross-aligned terms that act as filters for exploring portions of a large unfamiliar corpus of search results that might most closely align with their home domain concepts. In either approach, we could consider exposing relevant vocabulary terms in both source and target domains to guide natural user queries. 

In keeping with the richness of translation work as potentially navigating and integrating multiple related novel concepts from a new domain, rather than translating individual terms into their direct equivalents, we also wonder about alternative designs that operate over \textit{collections} of alignments, rather than individual lists of search results. Such designs may align better with our Case Study 2 participant's preference for making sense of collections of terms, rather than individual terms. 
In particular, the participant wanted to get a sense of all the response terms at once, before being able to assess how relevant or novel a term was. Judging in-silo versus all-at-once is certainly a design consideration. The former might lead to more criticality and premature rejection, which could lead to frustration if many returned terms are not relevant even if a minority are fruitful alignments to specialized concepts (akin to the precision versus recall tradeoff). The latter might give a user more of a mental sketch of grouping and clustering thematic conceptual mappings, better rewarding the experience given the presence of frutiful alignments, even if they are situated next to irrelevant concepts. The framing of the tool as a search interface might also lead users to desire more directly relevant terms or 1-to-1 mappings, given broader themes in HCI around how affordances of a search interface can substantially shape how people assess the relevance of cross-domain inspirations \cite{kangAugmentingScientificCreativity2022,srinivasanImprovingSelectionAnalogical2024}, or use/adapt them for novel insights \cite{chanFormulatingFixatingEffects2024b}.

Overall, we believe framing interdisciplinary foraging in terms of \textit{translation work}, in alignment with multilingual NLP, has strong potential to yield useful new designs for intelligent interactive systems that support cross-domain information-seeking.

\bibliographystyle{ACM-Reference-Format}
\bibliography{references, references_aclanthology}

\appendix


\section{Qualitative Assessments of Mappings}
\label{ap:qualitative_assessments}
\subsection{Assessments for Case Study 1}
\onecolumn
    \begin{longtable}{|l|l|l|l|l|}
    \hline
         \rowcolor{yellow!20} query\_term & cross\_term & cossim & relevance & novelty  \\ \hline 
        examples & gets & 0.6043121219 & 0 & 0 \\ \hline
        examples & gets & 0.6043121219 & 0 & 0 \\ \hline
        examples & gets & 0.6043121219 & 0 & 0 \\ \hline
        examples & gets & 0.6043121219 & 0 & 0 \\ \hline
        examples & gets & 0.6043121219 & 0 & 0 \\ \hline
        examples & substitutes & 0.5895254016 & 2 & 2 \\ \hline
        examples & substitutes & 0.5895254016 & 2 & 2 \\ \hline
        examples & substitutes & 0.5895254016 & 1 & 0 \\ \hline
        examples & substitutes & 0.5895254016 & 1 & 0 \\ \hline
        examples & substitutes & 0.5895254016 & 2 & 0 \\ \hline
        examples & notebook & 0.5892814994 & 0 & 0 \\ \hline
        examples & notebook & 0.5892814994 & 0 & 0 \\ \hline
        examples & pockets & 0.5892053246 & 2 & 2 \\ \hline
        examples & pockets & 0.5892053246 & 2 & 2 \\ \hline
        examples & welfare & 0.5747747421 & 0 & 0 \\ \hline
        examples & welfare & 0.5747747421 & 0 & 0 \\ \hline
        examples & welfare & 0.5747747421 & 0 & 0 \\ \hline
        examples & welfare & 0.5747747421 & 0 & 0 \\ \hline
        examples & welfare & 0.5747747421 & 0 & 0 \\ \hline
        examples & substitute & 0.5725717545 & 2 & 2 \\ \hline
        examples & substitute & 0.5725717545 & 2 & 2 \\ \hline
        examples & substitute & 0.5725717545 & 2 & 2 \\ \hline
        examples & substitute & 0.5725717545 & 2 & 2 \\ \hline
        examples & substitute & 0.5725717545 & 2 & 2 \\ \hline
        examples & unaware & 0.5722236037 & 0 & 0 \\ \hline
        examples & unaware & 0.5722236037 & 0 & 0 \\ \hline
        examples & unaware & 0.5722236037 & 0 & 0 \\ \hline
        examples & unaware & 0.5722236037 & 0 & 0 \\ \hline
        examples & unaware & 0.5722236037 & 0 & 0 \\ \hline
        examples & push & 0.5718544126 & 1 & 1 \\ \hline
        examples & push & 0.5718544126 & 1 & 1 \\ \hline
        examples & push & 0.5718544126 & 0 & 0 \\ \hline
        examples & push & 0.5718544126 & 1 & 1 \\ \hline
        examples & push & 0.5718544126 & 0 & 0 \\ \hline
        examples & offset & 0.5687605739 & 0 & 0 \\ \hline
        examples & offset & 0.5687605739 & 0 & 0 \\ \hline
        examples & offset & 0.5687605739 & 0 & 0 \\ \hline
        examples & offset & 0.5687605739 & 0 & 0 \\ \hline
        examples & else & 0.5636897683 & 0 & 0 \\ \hline
        examples & else & 0.5636897683 & 0 & 0 \\ \hline
        examples & else & 0.5636897683 & 0 & 0 \\ \hline
        examples & else & 0.5636897683 & 0 & 0 \\ \hline
        examples & else & 0.5636897683 & 0 & 0 \\ \hline
        examples & buyers & 0.5636130571 & 0 & 0 \\ \hline
        examples & buyers & 0.5636130571 & 0 & 0 \\ \hline
        examples & buyers & 0.5636130571 & 0 & 0 \\ \hline
        examples & buyers & 0.5636130571 & 0 & 0 \\ \hline
        examples & buyers & 0.5636130571 & 0 & 0 \\ \hline
        examples & everyone & 0.5618861914 & 0 & 0 \\ \hline
        stimulus & trajectories & 0.6212699413 & 2 & 2 \\ \hline
        stimulus & trajectories & 0.6212699413 & 2 & 2 \\ \hline
        stimulus & trajectories & 0.6212699413 & 2 & 2 \\ \hline
        stimulus & trajectories & 0.6212699413 & 2 & 2 \\ \hline
        stimulus & trajectories & 0.6212699413 & 2 & 2 \\ \hline
        stimulus & spillovers & 0.6189801693 & 2 & 2 \\ \hline
        stimulus & spillovers & 0.6189801693 & 2 & 2 \\ \hline
        stimulus & spillovers & 0.6189801693 & 2 & 2 \\ \hline
        stimulus & spillovers & 0.6189801693 & 2 & 2 \\ \hline
        stimulus & spillovers & 0.6189801693 & 2 & 2 \\ \hline
        stimulus & varies & 0.6030950546 & 2 & 2 \\ \hline
        stimulus & varies & 0.6030950546 & 1 & 1 \\ \hline
        stimulus & varies & 0.6030950546 & 1 & 0 \\ \hline
        stimulus & varies & 0.6030950546 & 0 & 0 \\ \hline
        stimulus & varies & 0.6030950546 & 0 & 0 \\ \hline
        stimulus & cultures & 0.5986304879 & 2 & 2 \\ \hline
        stimulus & cultures & 0.5986304879 & 1 & 2 \\ \hline
        stimulus & cultures & 0.5986304879 & 1 & 2 \\ \hline
        stimulus & cultures & 0.5986304879 & 1 & 2 \\ \hline
        stimulus & cultures & 0.5986304879 & 1 & 2 \\ \hline
        stimulus & domestic & 0.5788759589 & 1 & 2 \\ \hline
        stimulus & domestic & 0.5788759589 & 1 & 2 \\ \hline
        stimulus & domestic & 0.5788759589 & 1 & 2 \\ \hline
        stimulus & domestic & 0.5788759589 & 1 & 2 \\ \hline
        stimulus & domestic & 0.5788759589 & 1 & 2 \\ \hline
        stimulus & histories & 0.5694535375 & 2 & 2 \\ \hline
        stimulus & histories & 0.5694535375 & 2 & 2 \\ \hline
        stimulus & histories & 0.5694535375 & 2 & 2 \\ \hline
        stimulus & histories & 0.5694535375 & 1 & 1 \\ \hline
        stimulus & histories & 0.5694535375 & 1 & 1 \\ \hline
        stimulus & cultural & 0.563362658 & 2 & 2 \\ \hline
        stimulus & cultural & 0.563362658 & 1 & 2 \\ \hline
        stimulus & cultural & 0.563362658 & 1 & 2 \\ \hline
        stimulus & cultural & 0.563362658 & 1 & 2 \\ \hline
        stimulus & cultural & 0.563362658 & 1 & 2 \\ \hline
        stimulus & disputes & 0.561938107 & 1 & 1 \\ \hline
        stimulus & disputes & 0.561938107 & 1 & 1 \\ \hline
        stimulus & disputes & 0.561938107 & 1 & 2 \\ \hline
        stimulus & disputes & 0.561938107 & 0 & 0 \\ \hline
        stimulus & economies & 0.5614743233 & 2 & 2 \\ \hline
        stimulus & economies & 0.5614743233 & 1 & 1 \\ \hline
        stimulus & economies & 0.5614743233 & 1 & 1 \\ \hline
        stimulus & economies & 0.5614743233 & 0 & 0 \\ \hline
        stimulus & economies & 0.5614743233 & 0 & 0 \\ \hline
        stimulus & boundaries & 0.5581468344 & 2 & 1 \\ \hline
        stimulus & boundaries & 0.5581468344 & 1 & 2 \\ \hline
        stimulus & boundaries & 0.5581468344 & 1 & 2 \\ \hline
        stimulus & boundaries & 0.5581468344 & 1 & 2 \\ \hline
        stimulus & boundaries & 0.5581468344 & 1 & 2 \\ \hline
        stimulus & branches & 0.5577203035 & 2 & 2 \\ \hline
    \caption{Response terms for query terms of interest from Psychology $\rightarrow$ Management / Organizational Science, paired with cosine similarity and qualitative assessments from one annotator.}\label{tab:appendix_qualitative_assessment_study_1}
\end{longtable}
\subsection{Assessments for Case Study 2}
\onecolumn
\begin{longtable}  {|l|l|l|l|l|l|l|l|}
    \hline
     \rowcolor{yellow!20} source & target & query\_term & cross\_term & similarity & annotator1 & annotator2 & avg  \\\hline
        ling & hci & translanguaging & deliberative & 0.8743163943 & -1 & 0 & -0.5 \\ \hline
        ling & hci & translanguaging & mixed-initiative & 0.8533095717 & 1 & 1 & 1 \\ \hline
        ling & hci & translanguaging & initiative & 0.8460384011 & -1 & 0 & -0.5 \\ \hline
        ling & hci & translanguaging & cooperative & 0.8315920234 & 1 & 0 & 0.5 \\ \hline
        ling & hci & translanguaging & speculative & 0.8281762004 & -1 & -1 & -1 \\ \hline
        ling & hci & translanguaging & proactive & 0.8109608293 & -1 & -1 & -1 \\ \hline
        ling & hci & translanguaging & co-creative & 0.8079094887 & 1 & 1 & 1 \\ \hline
        ling & hci & translanguaging & Creative & 0.8072760105 & -1 & 0 & -0.5 \\ \hline
        ling & hci & translanguaging & adaptive & 0.8055086136 & -1 & 0 & -0.5 \\ \hline
        ling & edu & translanguaging & translanguaging & 0.790014565 & 0 & 1 & 0.5 \\ \hline
        ling & edu & translanguaging & teacher-talk & 0.784806788 & 1 & 1 & 1 \\ \hline
        ling & edu & translanguaging & transliteracy & 0.782607913 & 1 & 1 & 1 \\ \hline
        ling & edu & translanguaging & classroom-based & 0.7754539251 & -1 & -1 & -1 \\ \hline
        ling & edu & translanguaging & language-learning & 0.7680151463 & -1 & 0 & -0.5 \\ \hline
        ling & edu & translanguaging & scaffold & 0.7634933591 & -1 & 0 & -0.5 \\ \hline
        ling & edu & translanguaging & literacy & 0.7614750266 & -1 & 0.5 & -0.25 \\ \hline
        ling & edu & translanguaging & translingual & 0.7595765591 & 0 & 1 & 0.5 \\ \hline
        ling & edu & translanguaging & languaging & 0.7590286136 & 0 & 1 & 0.5 \\ \hline
        ling & edu & translanguaging & teaching/learning & 0.7577226758 & -1 & -1 & -1 \\ \hline
        ling & imm & translanguaging & adapts & 0.8113001585 & -1 & 0 & -0.5 \\ \hline
        ling & imm & translanguaging & machine & 0.7610801458 & -1 & -1 & -1 \\ \hline
        ling & imm & translanguaging & EBIs & 0.7520438433 & 1 & -1 & 0 \\ \hline
        ling & imm & translanguaging & mentoring & 0.7489615083 & -1 & -1 & -1 \\ \hline
        ling & imm & translanguaging & adaptations & 0.747436583 & -1 & 0 & -0.5 \\ \hline
        ling & imm & translanguaging & impetus & 0.7393995523 & -1 & -1 & -1 \\ \hline
        ling & imm & translanguaging & adaption & 0.7382391691 & -1 & 0 & -0.5 \\ \hline
        ling & imm & translanguaging & crosscultural & 0.7377848029 & 0 & 0 & 0 \\ \hline
        ling & imm & translanguaging & intercultural & 0.7375655174 & 0 & 0 & 0 \\ \hline
        ling & imm & translanguaging & multicultural & 0.7371763587 & 0 & 0 & 0 \\ \hline
        ling & eth & translanguaging & transcultural & 0.8437370062 & 1 & 1 & 1 \\ \hline
        ling & eth & translanguaging & cultural & 0.8020462394 & -1 & 0 & -0.5 \\ \hline
        ling & eth & translanguaging & cultural\_\_identity & 0.7983759642 & 1 & 1 & 1 \\ \hline
        ling & eth & translanguaging & EICM & 0.7838115692 & 1 & 1 & 1 \\ \hline
        ling & eth & translanguaging & multicultural\_\_education & 0.7721879482 & 0 & 1 & 0.5 \\ \hline
        ling & eth & translanguaging & bicultural\_\_identity & 0.771873951 & 1 & 1 & 1 \\ \hline
        ling & eth & translanguaging & ethno-cultural & 0.7650336623 & 1 & 0 & 0.5 \\ \hline
        ling & eth & translanguaging & multi-cultural & 0.7632349133 & 0 & 0 & 0 \\ \hline
        ling & eth & translanguaging & acculturation\_\_process & 0.7588860989 & 1 & 1 & 1 \\ \hline
        ling & eth & translanguaging & best\_\_practices & 0.757483542 & -1 & -1 & -1 \\ \hline
        ling & pro & translanguaging & CLIL & 0.8253515363 & 0 & 0 & 0 \\ \hline
        ling & pro & translanguaging & learning & 0.8214411736 & -1 & 0 & -0.5 \\ \hline
        ling & pro & translanguaging & situated\_\_learning & 0.8092653155 & 1 & 0 & 0.5 \\ \hline
        ling & pro & translanguaging & CID & 0.8076415658 & 1 & 1 & 1 \\ \hline
        ling & pro & translanguaging & pedagogic & 0.7999457121 & -1 & 0 & -0.5 \\ \hline
        ling & pro & translanguaging & M-learning & 0.7973175049 & -1 & 0 & -0.5 \\ \hline
        ling & pro & translanguaging & pedagogical & 0.7970983386 & -1 & 0 & -0.5 \\ \hline
        ling & pro & translanguaging & e-learning & 0.7967173457 & 0 & 0 & 0 \\ \hline
        ling & pro & translanguaging & student\_\_learning & 0.7891011238 & 0 & 0 & 0 \\ \hline
        ling & pro & translanguaging & situated\_\_learning\_\_theory & 0.7856515646 & 1 & 0 & 0.5 \\ \hline
        ling & hci & repertoire & vision & 0.6829593182 & -1 & -1 & -1 \\ \hline
        ling & hci & repertoire & participation & 0.6826735139 & -1 & 0 & -0.5 \\ \hline
        ling & hci & repertoire & Real-time & 0.674511373 & -1 & -1 & -1 \\ \hline
        ling & hci & repertoire & envision & 0.6718433499 & -1 & -1 & -1 \\ \hline
        ling & hci & repertoire & freedom & 0.6680535674 & -1 & -1 & -1 \\ \hline
        ling & hci & repertoire & path & 0.6620223522 & -1 & -1 & -1 \\ \hline
        ling & hci & repertoire & puppet & 0.6607789993 & -1 & -1 & -1 \\ \hline
        ling & hci & repertoire & self-control & 0.6590454578 & -1 & -1 & -1 \\ \hline
        ling & hci & repertoire & partner & 0.6556383371 & -1 & -1 & -1 \\ \hline
        ling & hci & repertoire & influential & 0.6525364518 & -1 & -1 & -1 \\ \hline
        ling & edu & repertoire & grounds & 0.6947515607 & -1 & 0 & -0.5 \\ \hline
        ling & edu & repertoire & flows & 0.6836689115 & -1 & 0 & -0.5 \\ \hline
        ling & edu & repertoire & unique & 0.6813056469 & -1 & -1 & -1 \\ \hline
        ling & edu & repertoire & repertoires & 0.6811887026 & 0 & 1 & 0.5 \\ \hline
        ling & edu & repertoire & repertoire & 0.6778647304 & 0 & 1 & 0.5 \\ \hline
        ling & edu & repertoire & resource & 0.6778272986 & -1 & 0 & -0.5 \\ \hline
        ling & edu & repertoire & accumulated & 0.6653056741 & -1 & 0 & -0.5 \\ \hline
        ling & edu & repertoire & reorient & 0.6639546752 & -1 & -1 & -1 \\ \hline
        ling & edu & repertoire & truths & 0.6492972374 & -1 & -1 & -1 \\ \hline
        ling & edu & repertoire & tacit & 0.6480041742 & 0 & 1 & 0.5 \\ \hline
        ling & imm & repertoire & new\_\_culture & 0.7574108243 & 0 & 0 & 0 \\ \hline
        ling & imm & repertoire & hence & 0.7198644876 & -1 & -1 & -1 \\ \hline
        ling & imm & repertoire & adapts & 0.7161327004 & -1 & 0 & -0.5 \\ \hline
        ling & imm & repertoire & successfully & 0.7103006244 & -1 & -1 & -1 \\ \hline
        ling & imm & repertoire & dominant\_\_culture & 0.7074056268 & 0 & -1 & -0.5 \\ \hline
        ling & imm & repertoire & s & 0.7032558918 & -1 & -1 & -1 \\ \hline
        ling & imm & repertoire & different\_\_cultures & 0.6992135644 & 0 & 0 & 0 \\ \hline
        ling & imm & repertoire & adaption & 0.6937709451 & -1 & 0 & -0.5 \\ \hline
        ling & imm & repertoire & host\_\_culture & 0.6921414733 & 0 & 1 & 0.5 \\ \hline
        ling & imm & repertoire & navigation & 0.6919963956 & -1 & 0 & -0.5 \\ \hline
        ling & eth & repertoire & self & 0.7625374794 & -1 & 0 & -0.5 \\ \hline
        ling & eth & repertoire & heritage\_\_culture & 0.751629293 & 1 & 1 & 1 \\ \hline
        ling & eth & repertoire & worldview & 0.7122991085 & 0 & 1 & 0.5 \\ \hline
        ling & eth & repertoire & sense & 0.711098969 & -1 & 0 & -0.5 \\ \hline
        ling & eth & repertoire & subculture & 0.7054489255 & 0 & 1 & 0.5 \\ \hline
        ling & eth & repertoire & tangible & 0.704145968 & -1 & -1 & -1 \\ \hline
        ling & eth & repertoire & heritage & 0.6999636889 & 0 & 1 & 0.5 \\ \hline
        ling & eth & repertoire & socio-cultural & 0.6926341653 & -1 & 0 & -0.5 \\ \hline
        ling & eth & repertoire & markers & 0.69101125 & -1 & 0 & -0.5 \\ \hline
        ling & eth & repertoire & belongs & 0.6903909445 & -1 & -1 & -1 \\ \hline
        ling & pro & repertoire & ideology & 0.7007747889 & -1 & 0 & -0.5 \\ \hline
        ling & pro & repertoire & repertoire & 0.6773253679 & 0 & 1 & 0.5 \\ \hline
        ling & pro & repertoire & lifeworld & 0.6770594716 & 1 & 1 & 1 \\ \hline
        ling & pro & repertoire & diﬀerent & 0.6749215126 & -1 & -1 & -1 \\ \hline
        ling & pro & repertoire & frames & 0.6726770997 & -1 & 0 & -0.5 \\ \hline
        ling & pro & repertoire & sensemaking & 0.6652895808 & 1 & 0 & 0.5 \\ \hline
        ling & pro & repertoire & competent & 0.6618661284 & -1 & 0 & -0.5 \\ \hline
        ling & pro & repertoire & manifest & 0.6616282463 & -1 & -1 & -1 \\ \hline
        ling & pro & repertoire & core & 0.6598476768 & -1 & 0 & -0.5 \\ \hline
        edu & hci & funds\_\_of\_\_knowledge & platform & 0.781198144 & -1 & 0 & -0.5 \\ \hline
        edu & hci & funds\_\_of\_\_knowledge & platforms & 0.7583290339 & -1 & 0 & -0.5 \\ \hline
        edu & hci & funds\_\_of\_\_knowledge & forms & 0.7142107487 & -1 & -1 & -1 \\ \hline
        edu & hci & funds\_\_of\_\_knowledge & formats & 0.7071314454 & -1 & -1 & -1 \\ \hline
        edu & hci & funds\_\_of\_\_knowledge & links & 0.7066740394 & -1 & 0 & -0.5 \\ \hline
        edu & hci & funds\_\_of\_\_knowledge & norm & 0.7046322227 & -1 & 0 & -0.5 \\ \hline
        edu & hci & funds\_\_of\_\_knowledge & grief & 0.6988732815 & -1 & -1 & -1 \\ \hline
        edu & hci & funds\_\_of\_\_knowledge & format & 0.697663188 & -1 & -1 & -1 \\ \hline
        edu & hci & funds\_\_of\_\_knowledge & Algorithms & 0.6881314516 & 0 & 0 & 0 \\ \hline
        edu & hci & funds\_\_of\_\_knowledge & theoretically & 0.6832361221 & -1 & 0 & -0.5 \\ \hline
        edu & ling & funds\_\_of\_\_knowledge & theorize & 0.7694742084 & -1 & 0 & -0.5 \\ \hline
        edu & ling & funds\_\_of\_\_knowledge & knowledge-rich & 0.7588395476 & -1 & 0 & -0.5 \\ \hline
        edu & ling & funds\_\_of\_\_knowledge & conceptualized & 0.7509846687 & -1 & 0 & -0.5 \\ \hline
        edu & ling & funds\_\_of\_\_knowledge & lens & 0.7487373352 & -1 & 0 & -0.5 \\ \hline
        edu & ling & funds\_\_of\_\_knowledge & conceptualize & 0.747408092 & -1 & 0 & -0.5 \\ \hline
        edu & ling & funds\_\_of\_\_knowledge & knowledges & 0.725109756 & -1 & 1 & 0 \\ \hline
        edu & ling & funds\_\_of\_\_knowledge & deploys & 0.7211380601 & -1 & -1 & -1 \\ \hline
        edu & ling & funds\_\_of\_\_knowledge & knowledge & 0.7191357613 & -1 & 0 & -0.5 \\ \hline
        edu & ling & funds\_\_of\_\_knowledge & multimodality & 0.711112082 & 0 & 1 & 0.5 \\ \hline
        edu & imm & funds\_\_of\_\_knowledge & conceptualize & 0.7420441508 & -1 & 0 & -0.5 \\ \hline
        edu & imm & funds\_\_of\_\_knowledge & contextualize & 0.730682373 & -1 & 0 & -0.5 \\ \hline
        edu & imm & funds\_\_of\_\_knowledge & contextualizes & 0.7160271406 & -1 & 0 & -0.5 \\ \hline
        edu & imm & funds\_\_of\_\_knowledge & multidimensional & 0.7149877548 & 0 & 0 & 0 \\ \hline
        edu & imm & funds\_\_of\_\_knowledge & acculturation\_\_theory & 0.7143118382 & 1 & -1 & 0 \\ \hline
        edu & imm & funds\_\_of\_\_knowledge & conceptual\_\_framework & 0.714176178 & 1 & 0 & 0.5 \\ \hline
        edu & imm & funds\_\_of\_\_knowledge & tridimensional & 0.7110076547 & 1 & 0 & 0.5 \\ \hline
        edu & imm & funds\_\_of\_\_knowledge & conceptualizes & 0.7109881043 & -1 & 0 & -0.5 \\ \hline
        edu & imm & funds\_\_of\_\_knowledge & conceptual\_\_model & 0.7070842385 & -1 & 0 & -0.5 \\ \hline
        edu & imm & funds\_\_of\_\_knowledge & multidimensionality & 0.7054414153 & -1 & 0 & -0.5 \\ \hline
        edu & eth & funds\_\_of\_\_knowledge & derives & 0.7214701772 & -1 & -1 & -1 \\ \hline
        edu & eth & funds\_\_of\_\_knowledge & conceptualize & 0.7162780762 & -1 & 0 & -0.5 \\ \hline
        edu & eth & funds\_\_of\_\_knowledge & theoretical\_\_approaches & 0.7159905434 & -1 & 0 & -0.5 \\ \hline
        edu & eth & funds\_\_of\_\_knowledge & synthesize & 0.7152888179 & -1 & 0 & -0.5 \\ \hline
        edu & eth & funds\_\_of\_\_knowledge & group\_\_identity & 0.7140477896 & 1 & 1 & 1 \\ \hline
        edu & eth & funds\_\_of\_\_knowledge & derived & 0.7053064108 & -1 & -1 & -1 \\ \hline
        edu & eth & funds\_\_of\_\_knowledge & existing & 0.7047510743 & -1 & -1 & -1 \\ \hline
        edu & eth & funds\_\_of\_\_knowledge & bridge & 0.7008541226 & -1 & 0 & -0.5 \\ \hline
        edu & eth & funds\_\_of\_\_knowledge & theorize & 0.6947138906 & -1 & 0 & -0.5 \\ \hline
        edu & eth & funds\_\_of\_\_knowledge & conceptualized & 0.6915852427 & 0 & 0 & 0 \\ \hline
        edu & pro & funds\_\_of\_\_knowledge & conceptualised & 0.7170065045 & 0 & 0 & 0 \\ \hline
        edu & pro & funds\_\_of\_\_knowledge & concepts & 0.7028673291 & 0 & 0 & 0 \\ \hline
        edu & pro & funds\_\_of\_\_knowledge & conceptualize & 0.7015826106 & 0 & 0 & 0 \\ \hline
        edu & pro & funds\_\_of\_\_knowledge & integrate & 0.6972215772 & 0 & 0 & 0 \\ \hline
        edu & pro & funds\_\_of\_\_knowledge & concept & 0.6937665939 & -1 & 0 & -0.5 \\ \hline
        edu & pro & funds\_\_of\_\_knowledge & habitus & 0.6908137798 & 1 & 1 & 1 \\ \hline
        edu & pro & funds\_\_of\_\_knowledge & conceptualise & 0.6904374957 & -1 & 0 & -0.5 \\ \hline
        edu & pro & funds\_\_of\_\_knowledge & understandings & 0.6889657378 & -1 & 0 & -0.5 \\ \hline
        edu & pro & funds\_\_of\_\_knowledge & theory & 0.6889147758 & -1 & 0 & -0.5 \\ \hline
        edu & hci & third\_\_space & insights & 0.7838318944 & -1 & -1 & -1 \\ \hline
        edu & hci & third\_\_space & intersectional & 0.7603951097 & 0 & 1 & 0.5 \\ \hline
        edu & hci & third\_\_space & lights & 0.7572006583 & -1 & -1 & -1 \\ \hline
        edu & hci & third\_\_space & CBE & 0.7424249053 & 0 & -1 & -0.5 \\ \hline
        edu & hci & third\_\_space & future & 0.7412995696 & -1 & -1 & -1 \\ \hline
        edu & hci & third\_\_space & puppet & 0.7375090122 & -1 & -1 & -1 \\ \hline
        edu & hci & third\_\_space & instructors & 0.734104991 & -1 & -1 & -1 \\ \hline
        edu & hci & third\_\_space & interdisciplinary & 0.7302479744 & 0 & 0 & 0 \\ \hline
        edu & hci & third\_\_space & discuss & 0.728939414 & -1 & -1 & -1 \\ \hline
        edu & ling & third\_\_space & multi/plurilingual & 0.7795246243 & 1 & 1 & 1 \\ \hline
        edu & ling & third\_\_space & continua & 0.766535759 & 1 & 0 & 0.5 \\ \hline
        edu & ling & third\_\_space & readership & 0.7573106289 & -1 & -1 & -1 \\ \hline
        edu & ling & third\_\_space & reconceptualise & 0.7523953319 & -1 & 0 & -0.5 \\ \hline
        edu & ling & third\_\_space & intertextuality & 0.7505890131 & 1 & 1 & 1 \\ \hline
        edu & ling & third\_\_space & conceptualized & 0.7432819605 & -1 & -1 & -1 \\ \hline
        edu & ling & third\_\_space & concept & 0.7429624796 & -1 & -1 & -1 \\ \hline
        edu & ling & third\_\_space & reconceptualization & 0.7417594194 & 1 & -1 & 0 \\ \hline
        edu & ling & third\_\_space & conceptualizes & 0.7408658266 & -1 & -1 & -1 \\ \hline
        edu & ling & third\_\_space & nexus & 0.7386583686 & -1 & 0 & -0.5 \\ \hline
        edu & imm & third\_\_space & TDE & 0.7607435584 & 1 & 1 & 1 \\ \hline
        edu & imm & third\_\_space & construction & 0.7474842072 & -1 & -1 & -1 \\ \hline
        edu & imm & third\_\_space & theorised & 0.7378210425 & -1 & -1 & -1 \\ \hline
        edu & imm & third\_\_space & concept & 0.7370481491 & -1 & -1 & -1 \\ \hline
        edu & imm & third\_\_space & conceptual\_\_framework & 0.728639245 & 1 & -1 & 0 \\ \hline
        edu & imm & third\_\_space & proposal & 0.7261104584 & -1 & -1 & -1 \\ \hline
        edu & imm & third\_\_space & information\_\_worlds & 0.7233423591 & 1 & 1 & 1 \\ \hline
        edu & imm & third\_\_space & enrich & 0.7202795744 & -1 & -1 & -1 \\ \hline
        edu & imm & third\_\_space & acculturation\_\_theory & 0.7174776196 & 0 & 0 & 0 \\ \hline
        edu & imm & third\_\_space & reconstruction & 0.716917634 & 1 & -1 & 0 \\ \hline
        edu & eth & third\_\_space & identity. & 0.7945914865 & -1 & 0 & -0.5 \\ \hline
        edu & eth & third\_\_space & identity\_\_construction & 0.7858144045 & 0 & 0 & 0 \\ \hline
        edu & eth & third\_\_space & social\_\_identity\_\_theories & 0.773149848 & 1 & 0 & 0.5 \\ \hline
        edu & eth & third\_\_space & social\_\_identity\_\_theory & 0.7703753114 & 1 & 0 & 0.5 \\ \hline
        edu & eth & third\_\_space & social\_\_identity\_\_approach & 0.7689462304 & 1 & 0 & 0.5 \\ \hline
        edu & eth & third\_\_space & identity\_\_theory & 0.7616072297 & 0 & 0 & 0 \\ \hline
        edu & eth & third\_\_space & constructionist & 0.7581090927 & 1 & 0 & 0.5 \\ \hline
        edu & eth & third\_\_space & hybrid & 0.7478024364 & -1 & 1 & 0 \\ \hline
        edu & eth & third\_\_space & identity\_\_negotiation & 0.7414662242 & 1 & 0 & 0.5 \\ \hline
        edu & eth & third\_\_space & social\_\_identity & 0.7385947704 & 0 & 0 & 0 \\ \hline
        edu & pro & third\_\_space & CoPs & 0.7541846633 & 1 & 1 & 1 \\ \hline
        edu & pro & third\_\_space & entrepreneurship & 0.7500813603 & -1 & 0 & -0.5 \\ \hline
        edu & pro & third\_\_space & research-practice & 0.7392860055 & -1 & 1 & 0 \\ \hline
        edu & pro & third\_\_space & conceptualized & 0.7345022559 & -1 & -1 & -1 \\ \hline
        edu & pro & third\_\_space & distinctive & 0.7331488132 & -1 & -1 & -1 \\ \hline
        edu & pro & third\_\_space & conceptualised & 0.7331167459 & -1 & -1 & -1 \\ \hline
        edu & pro & third\_\_space & brokerage & 0.7313123345 & 1 & 1 & 1 \\ \hline
        edu & pro & third\_\_space & conceptualise & 0.7266747952 & -1 & -1 & -1 \\ \hline
        edu & pro & third\_\_space & boundary-crossing & 0.7237938046 & 1 & 1 & 1 \\ \hline
        edu & pro & third\_\_space & brokering & 0.7229532003 & 1 & 1 & 1 \\ \hline
        hci & ling & AI-mediated\_\_communication & transnational & 0.8087528944 & 1 & 1 & 1 \\ \hline
        hci & ling & AI-mediated\_\_communication & transnationalism & 0.8066139817 & 1 & 1 & 1 \\ \hline
        hci & ling & AI-mediated\_\_communication & Everyday & 0.7917954922 & -1 & -1 & -1 \\ \hline
        hci & ling & AI-mediated\_\_communication & Transnational & 0.7847146392 & -1 & 1 & 0 \\ \hline
        hci & ling & AI-mediated\_\_communication & Looking & 0.7837681174 & -1 & -1 & -1 \\ \hline
        hci & ling & AI-mediated\_\_communication & racialization & 0.78137362 & 0 & -1 & -0.5 \\ \hline
        hci & ling & AI-mediated\_\_communication & emergency & 0.7741872072 & -1 & -1 & -1 \\ \hline
        hci & ling & AI-mediated\_\_communication & Masculinities & 0.7690383196 & -1 & -1 & -1 \\ \hline
        hci & ling & AI-mediated\_\_communication & Identities & 0.7651382089 & -1 & -1 & -1 \\ \hline
        hci & edu & AI-mediated\_\_communication & makerspaces & 0.77935642 & 0 & -1 & -0.5 \\ \hline
        hci & edu & AI-mediated\_\_communication & Partnerships & 0.7601863146 & -1 & -1 & -1 \\ \hline
        hci & edu & AI-mediated\_\_communication & Inclusive & 0.7550260425 & -1 & -1 & -1 \\ \hline
        hci & edu & AI-mediated\_\_communication & Immigrant & 0.7508330941 & -1 & -1 & -1 \\ \hline
        hci & edu & AI-mediated\_\_communication & interrogates & 0.7342031002 & -1 & -1 & -1 \\ \hline
        hci & edu & AI-mediated\_\_communication & Tongan & 0.7277263999 & -1 & -1 & -1 \\ \hline
        hci & edu & AI-mediated\_\_communication & anti-immigrant & 0.7198304534 & -1 & -1 & -1 \\ \hline
        hci & edu & AI-mediated\_\_communication & nepantla & 0.7160634995 & 1 & 0 & 0.5 \\ \hline
        hci & edu & AI-mediated\_\_communication & compassionate & 0.7143685222 & -1 & -1 & -1 \\ \hline
        hci & edu & AI-mediated\_\_communication & Collaboration & 0.71210289 & -1 & 0 & -0.5 \\ \hline
        hci & imm & AI-mediated\_\_communication & placemaking & 0.7857161164 & 1 & -1 & 0 \\ \hline
        hci & imm & AI-mediated\_\_communication & place-making & 0.7733535171 & 1 & ~ & 1 \\ \hline
        hci & imm & AI-mediated\_\_communication & HRM & 0.7627471685 & -1 & -1 & -1 \\ \hline
        hci & imm & AI-mediated\_\_communication & Divakaruni & 0.7560439706 & -1 & -1 & -1 \\ \hline
        hci & imm & AI-mediated\_\_communication & e-diaspora & 0.7536911964 & 1 & 0 & 0.5 \\ \hline
        hci & imm & AI-mediated\_\_communication & globalisation & 0.7535432577 & -1 & 0 & -0.5 \\ \hline
        hci & imm & AI-mediated\_\_communication & facilitating & 0.7498921752 & -1 & 0 & -0.5 \\ \hline
        hci & imm & AI-mediated\_\_communication & TDE & 0.7489527464 & 0 & 0 & 0 \\ \hline
        hci & imm & AI-mediated\_\_communication & talent & 0.7473618984 & -1 & -1 & -1 \\ \hline
        hci & imm & AI-mediated\_\_communication & advent & 0.7453843355 & -1 & -1 & -1 \\ \hline
        hci & eth & AI-mediated\_\_communication & wider & 0.8147138357 & -1 & -1 & -1 \\ \hline
        hci & eth & AI-mediated\_\_communication & Chagrin & 0.8083021641 & -1 & -1 & -1 \\ \hline
        hci & eth & AI-mediated\_\_communication & Falls & 0.808240056 & -1 & -1 & -1 \\ \hline
        hci & eth & AI-mediated\_\_communication & landscape & 0.7854062319 & -1 & -1 & -1 \\ \hline
        hci & eth & AI-mediated\_\_communication & landmark & 0.7750949264 & -1 & -1 & -1 \\ \hline
        hci & eth & AI-mediated\_\_communication & agenda & 0.7665138245 & -1 & -1 & -1 \\ \hline
        hci & eth & AI-mediated\_\_communication & BAM & 0.7627230883 & -1 & -1 & -1 \\ \hline
        hci & eth & AI-mediated\_\_communication & scientists & 0.7586004138 & -1 & -1 & -1 \\ \hline
        hci & eth & AI-mediated\_\_communication & institutional & 0.7544520497 & -1 & -1 & -1 \\ \hline
        hci & eth & AI-mediated\_\_communication & higher\_\_education & 0.7536155581 & -1 & -1 & -1 \\ \hline
        hci & pro & AI-mediated\_\_communication & APRT & 0.7847443819 & -1 & -1 & -1 \\ \hline
        hci & pro & AI-mediated\_\_communication & VAC & 0.763710022 & -1 & -1 & -1 \\ \hline
        hci & pro & AI-mediated\_\_communication & SGM/DSD & 0.7588301301 & -1 & -1 & -1 \\ \hline
        hci & pro & AI-mediated\_\_communication & Brazil & 0.7540581226 & -1 & -1 & -1 \\ \hline
        hci & pro & AI-mediated\_\_communication & BGI & 0.7479191422 & -1 & -1 & -1 \\ \hline
        hci & pro & AI-mediated\_\_communication & RCMI & 0.7463548183 & -1 & -1 & -1 \\ \hline
        hci & pro & AI-mediated\_\_communication & SBE & 0.7405706048 & 1 & -1 & 0 \\ \hline
        hci & pro & AI-mediated\_\_communication & GEM‐SLP & 0.7375261188 & -1 & -1 & -1 \\ \hline
        hci & pro & AI-mediated\_\_communication & wellbeing & 0.7351579666 & 0 & 0 & 0 \\ \hline
        hci & pro & AI-mediated\_\_communication & Delivery & 0.7338174582 & -1 & 0 & -0.5 \\ \hline
   \caption{Response terms for query terms of interest, paired with cosine similarity and qualitative assessments from two annotators.}\label{tab:appendix_qualitative_assessment}
\end{longtable}

\end{document}


\begin{center}
\textbf{Supplemental Materials}
\end{center}

\section{Seed Papers for Constructing Community Corpora for Case Study 1}
\label{ap:seed-papers}

\subsection{Psychology}
\begin{enumerate}
    \item \small Marsh, R. L., Landau, J. D., \& Hicks, J. L. (1996). How examples may (and may not) constrain creativity. Memory \& Cognition, 24(5), 669–680.
    \item \small Ward, T. B. (1994). Structured imagination: The role of category structure in exemplar generation. Cognitive Psychology, 27(1), 1–40. 
    \item \small Nijstad, B. A., \& Stroebe, W. (2006). How the group affects the mind: A cognitive model of idea generation in groups. Personality and Social Psychology Review, 10(3), 186–213. 
    \item \small Berg, J. M. (2014). The primal mark: How the beginning shapes the end in the development of creative ideas. Organizational Behavior and Human Decision Processes, 125(1), 1–17. 
    \item \small Ezzat, H., Agogué, M., Le Masson, P., Weil, B., \& Cassotti, M. (2020). Specificity and Abstraction of Examples: Opposite Effects on Fixation for Creative Ideation. The Journal of Creative Behavior, 54(1), 115–122. 
    \item \small Chrysikou, E. G., \& Weisberg, R. W. (2005). Following the wrong footsteps: Fixation effects of pictorial examples in a design problem-solving task. Journal of Experimental Psychology: Learning, Memory, and Cognition, 31(5), 1134–1148. 
    \item \small Kohn, N. W., \& Smith, S. M. (2011). Collaborative fixation: Effects of others’ ideas on brainstorming. Applied Cognitive Psychology, 25(3), 359–371.
\end{enumerate}

\subsection{Management / Organization Science}
\begin{enumerate}
    \item \small Kavadias, S., \& Sommer, S. C. . C. (2009). The Effects of Problem Structure and Team Diversity on Brainstorming Effectiveness. Management Science, 55(12), 1899–1913.
    \item \small Kneeland, M. K., Schilling, M. A., \& Aharonson, B. S. (2020). Exploring Uncharted Territory: Knowledge Search Processes in the Origination of Outlier Innovation. Organization Science, 31(3), 535–557.
    \item \small Althuizen, N., \& Wierenga, B. (2014). Supporting Creative Problem Solving with a Case-Based Reasoning System. Journal of Management Information Systems, 31(1), 309–340. 
    \item \small Csaszar, F. A., \& Siggelkow, N. (2010). How Much to Copy? Determinants of Effective Imitation Breadth. Organization Science, 21(3), 661–676. 
    \item \small Franke, N., Poetz, M. K., \& Schreier, M. (2014). Integrating Problem Solvers from Analogous Markets in New Product Ideation. Management Science, 60(4), 1063–1081.
    \item \small Mastrogiorgio, M., \& Gilsing, V. (2016). Innovation through exaptation and its determinants: The role of technological complexity, analogy making \& patent scope. Research Policy, 45(7), 1419–1435. 
    \item \small Andriani, P., \& Carignani, G. (2014). Modular exaptation: A missing link in the synthesis of artificial form. Research Policy, 43(9), 1608–1620. 
    \item \small Enkel, E., \& Gassmann, O. (2010). Creative imitation: Exploring the case of cross-industry innovation. R \& D Management, 40(3), 256–270.
\end{enumerate}

\section{Seed Papers for Constructing Community Corpora for Case Study B}\label{appendix:case_study_b_seed_papers}
\subsection{Applied Linguistics}

\begin{enumerate}
    \item \small Wei Li, Hua Zhu (2013). Translanguaging Identities and Ideologies: Creating Transnational Space Through Flexible Multilingual Practices Amongst Chinese University Students in the UK. Applied Linguistics 34,516-535 
    \item \small Melinda Martin‐Beltrán (2014). “What Do You Want to Say?” How Adolescents Use Translanguaging to Expand Learning Opportunities. International Multilingual Research Journal 8,208 - 230 
    \item \small Ricardo Otheguy, Ofelia García, Wallis Reid (2015). Clarifying translanguaging and deconstructing named languages: A perspective from linguistics. Applied Linguistics Review 6,281 - 307 
    \item \small Brooke Schreiber (2015). "I AM WHAT I AM": MULTILINGUAL IDENTITY AND DIGITAL TRANSLANGUAGING. Language Learning \& Technology 19,69-87 
    \item \small Li Wei (2011). Moment analysis and translanguaging space: discursive construction of identities by multilingual Chinese youth in Britain. Journal of Pragmatics 43,1222-1235 
    \item \small Li Wei (2018). Translanguaging as a Practical Theory of Language.  
    \item \small F. Grosjean (1985). The bilingual as a competent but specific speaker‐hearer. Journal of Multilingual and Multicultural Development 6,467-477 
    \item \small F. Grosjean (1989). Neurolinguists, beware! The bilingual is not two monolinguals in one person. Brain and Language 36,3-15 
    \item \small F. Grosjean (2001). The Bilingual's Language Modes..   
    \item \small F. Grosjean, Joanne L. Miller (1994). Going in and out of Languages: An Example of Bilingual Flexibility. Psychological Science 5,201 - 206 
    \item \small B. Busch (2012). The Linguistic Repertoire Revisited. Applied Linguistics 33,503-523 
    \item \small B. Busch (2015). Expanding the Notion of the Linguistic Repertoire: On the Concept of Spracherleben—The Lived Experience of Language. Applied Linguistics 38,340-358 
    \item \small M. Wetherell (1998). Positioning and Interpretative Repertoires: Conversation Analysis and Post-Structuralism in Dialogue. Discourse \& Society 9,387 - 412 
    \item \small Suzanne García Mateus (2014). Translanguaging: Language, Bilingualism, and Education. Bilingual Research Journal 37,366 - 369 
    \item \small Li Wei, Ofelia García (2022). Not a First Language but One Repertoire: Translanguaging as a Decolonizing Project. RELC Journal 53,313 - 324 
    \item \small Peijian Paul Sun, Lawrence Jun Zhang (2022). Effects of Translanguaging in Online Peer Feedback on Chinese University English-as-a-Foreign-Language Students’ Writing Performance. RELC Journal 53,325 - 341 
    \item \small J. Cenoz, D. Gorter (2022). Pedagogical Translanguaging and Its Application to Language Classes. RELC Journal 53,342 - 354 
    \item \small J. Cenoz, D. Gorter (2021). Pedagogical Translanguaging.  
    \item \small Li Wei (2017). Translanguaging as a Practical Theory of Language.  
    \item \small F. Fang, L. Zhang, Pramod K. Sah (2022). Translanguaging in Language Teaching and Learning: Current Practices and Future Directions. RELC Journal 53,305 - 312 
    \item \small Pramod K. Sah, Ryuko Kubota (2022). Towards critical translanguaging: a review of literature on English as a medium of instruction in South Asia’s school education. Asian Englishes 24,132 - 146 
    \item \small J. MacSwan (2017). A Multilingual Perspective on Translanguaging. American Educational Research Journal 54,167 - 201 
    \item \small Ofelia García, Ricardo Otheguy (2020). Plurilingualism and translanguaging: commonalities and divergences. International Journal of Bilingual Education and Bilingualism 23,17 - 35 
    \item \small A. Bell (1984). Language style as audience design. Language in Society 13,145 - 204 
    \item \small Si On Yoon, Sarah Brown-Schmidt (2019). Contextual Integration in Multiparty Audience Design. Cognitive science 43 12,e12807 
    \item \small Kumiko Fukumura (2015). Interface of Linguistic and Visual Information During Audience Design. Cognitive science 39 6,1419-33 
    \item \small W. Horton (2009). Memory and Other Limits on Audience Design in Reference Production.  
    \item \small V. Youssef (1993). Children's linguistic choices: Audience design and societal norms. Language in Society 22,257 - 274 
\end{enumerate}

\subsection{Education}

\begin{enumerate}
    \item \small Kris D. Gutiérrez (2008). Developing a Sociocritical Literacy in the Third Space. Reading Research Quarterly 43,148-164 
    \item \small Melinda Martin‐Beltrán (2010). Positioning proficiency: How students and teachers (de)construct language proficiency at school. Linguistics and Education 21,257-281 
    \item \small N. Mercer, C. Howe (2012). Explaining the dialogic processes of teaching and learning: The value and potential of sociocultural theory. Learning, Culture and Social Interaction 1,12-21 
    \item \small Kongji Qin, Faythe P Beauchemin (2022). “I Can Go Slapsticks”: Humor as Humanizing Pedagogy for Science Instruction With Multilingual Adolescent Immigrant Learners. Literacy Research: Theory, Method, and Practice 71,304 - 322 
    \item \small Ben Gleason (2020). Expanding interaction in online courses: integrating critical humanizing pedagogy for learner success. Educational Technology Research and Development 69,51 - 54 
    \item \small Dorinda J. Carter Andrews, Tasha M. Brown, Bernadette M. Castillo, Davena Jackson, V. Vellanki (2019). Beyond Damage-Centered Teacher Education: Humanizing Pedagogy for Teacher Educators and Preservice Teachers. Teachers College Record: The Voice of Scholarship in Education 121,1 - 28 
    \item \small Lori Czop Assaf, Kristie O’Donnell Lussier, Shelly Furness, Meagan A. Hoff (2019). Exploring How a Study Abroad and International Service-Learning Project Shaped Preservice Teachers’ Understanding of Humanizing Pedagogy. The Teacher Educator 54,105 - 124 
    \item \small María E. Fránquiz, Alba A. Ortiz, G. Lara (2019). Co-editor’s introduction: Humanizing pedagogy, research and learning. Bilingual Research Journal 42,381 - 386 
    \item \small S. Osorio (2018). Toward a humanizing pedagogy: Using Latinx children’s literature with early childhood students. Bilingual Research Journal 41,22 - 5 
    \item \small Margarita Zisselsberger (2016). Toward a humanizing pedagogy: Leveling the cultural and linguistic capital in a fifth-grade writing classroom. Bilingual Research Journal 39,121 - 137 
    \item \small Caroline Khene (2014). Supporting a Humanizing Pedagogy in the Supervision Relationship and Process: A Reflection in a Developing Country. International Journal of Doctoral Studies 9,073-083 
    \item \small L. Bartolomé (1994). Beyond the Methods Fetish: Toward a Humanizing Pedagogy. Harvard Educational Review 64,173-195 
    \item \small Teresa M. Huerta (2011). Humanizing Pedagogy: Beliefs and Practices on the Teaching of Latino Children. Bilingual Research Journal 34,38 - 57 
    \item \small Jennifer Counts, Antonis Katsiyannis, Denise K. Whitford (2018). Culturally and Linguistically Diverse Learners in Special Education: English Learners. NASSP Bulletin 102,21 - 5 
    \item \small Alpana Bhattacharya (2020). Preparing Preservice Teacher Candidates for Educating Culturally and Linguistically Diverse Learners.  ,76-103 
    \item \small Xenia Hadjioannou, M. Hutchinson, Marisa Hockman (2016). Addressing the Needs of 21st-Century Teachers Working with Culturally and Linguistically Diverse Learners.. The CATESOL Journal 28,1-29 
    \item \small M. Peercy (2016). New Pedagogies in Teacher Education for Teaching Linguistically Diverse Learners. International Multilingual Research Journal 10,165 - 167 
    \item \small Luciana Oliveira, Steven Z. Athanases (2017). A Framework to Reenvision Instructional Scaffolding for Linguistically Diverse Learners.. Journal of Adolescent \& Adult Literacy 61,123-129 
    \item \small R. M. Greene (2017). Gifted Culturally Linguistically Diverse Learners: A School-Based Exploration.   
    \item \small Roben W. Taylor, Alex Kumi-Yeboah, R. Ringlaben (2016). Pre-Service Teachers' Perceptions towards Multicultural Education \& Teaching of Culturally \& Linguistically Diverse Learners.. Multicultural Education 23,42-48 
    \item \small Thea Yurkewecz (2014). Observational Tools to Inform Instruction for Culturally and Linguistically Diverse Learners..  24,26-35 
    \item \small C. Lidz, Sheila L. Macrine (2001). An Alternative Approach to the Identification of Gifted Culturally and Linguistically Diverse Learners. School Psychology International 22,74 - 96 
    \item \small Ellen Strickland Kaje (2009). More than just good teaching: Teachers engaging culturally and linguistically diverse learners with content and language in mainstream classrooms.   
    \item \small N. Ruppert, Bridget K. Coleman, H. Pinter, Denise T. Johnson, Meghan A. Rector, Chandra C. Díaz (2022). Culturally Sustaining Practices for Middle Level Mathematics Teachers. Education Sciences 
    \item \small Chandra C. Díaz (2023). Culturally sustaining practices for middle level teacher educators. Middle School Journal 54,39 - 45 
    \item \small Rawn Boulden, Emily C. Goodman-Scott (2023). A Quantitative Exploration of School Counselors’ Evidence-Based Classroom Management Implementation: Investigating Culturally Sustaining Practices and Multicultural Competence. Professional School Counseling 27 
    \item \small Amy Vinlove (2017). Teach What You Know: Cultivating Culturally Sustaining Practices in Pre-Service Alaska Native Teachers.  29 
    \item \small K. Peppler, M. Dahn, Mizuko Ito (2022). Connected Arts Learning: Cultivating Equity Through Connected and Creative Educational Experiences. Review of Research in Education 46,264 - 287 
    \item \small Christopher P. Brown, D. Ku, K. Puckett, David P. Barry (2022). Preservice teachers’ struggles in finding culturally sustaining spaces in standardized teaching contexts. Journal of Early Childhood Teacher Education 44,463 - 483 
    \item \small Kelly K. Wissman (2021). Bringing a Culturally Sustaining Lens to Reading Intervention. Journal of Literacy Research 53,563 - 587 
    \item \small Christina J. Cavallaro, Sabrina F. Sembiante (2020). Facilitating culturally sustaining, functional literacy practices in a middle school ESOL reading program: a design-based research study. Language and Education 35,160 - 179 
    \item \small Sarah N. Newcomer (2020). “Who We Are Today”: Latinx Youth Perspectives on the Possibilities of Being Bilingual and Bicultural. Journal of Language, Identity \& Education 19,193 - 207 
    \item \small Jamila M. Kareem (2020). Chapter 16. Sustained Communities for Sustained Learning: Connecting Culturally Sustaining Pedagogy to WAC Learning Outcomes.  ,293-308 
    \item \small Katherine Wimpenny, K. Finardi, Marina Orsini-Jones, L. Jacobs (2022). Knowing, Being, Relating and Expressing Through Third Space Global South-North COIL: Digital Inclusion and Equity in International Higher Education. Journal of Studies in International Education 26,279 - 296 
    \item \small Cameron Smith, M. Holden, Eustacia Yu, Patrick Hanlon (2021). ‘So what do you do?’: Third space professionals navigating a Canadian university context. Journal of Higher Education Policy and Management 43,505 - 519 
    \item \small Harald Raaijmakers, Birgitta Mc Ewen, S. Walan, Nina Christenson (2021). Developing museum-school partnerships: art-based exploration of science issues in a third space. International Journal of Science Education 43,2746 - 2768 
    \item \small Jihea Maddamsetti (2021). Elementary ESL teachers’ advocacy for emerging bilinguals: a third space perspective. Pedagogies: An International Journal 18,153 - 181 
    \item \small Hongyu Wang, J. Flory (2021). Curriculum in a Third Space. Oxford Research Encyclopedia of Education 
    \item \small B. Ford (2021). From a Different Place to a Third Space: Rethinking International Student Pedagogy in the Western Conservatoire. The Politics of Diversity in Music Education 
    \item \small Marilee Coles-Ritchie, Walkie Charles (2022). Indigenizing Assessment Using Community Funds of Knowledge: A Critical Action Research Study. Journal of American Indian Education 50,26 - 41 
    \item \small Jodie Hunter (2022). Challenging and disrupting deficit discourses in mathematics education: positioning young diverse learners to document and share their mathematical funds of knowledge. Research in Mathematics Education 24,187 - 201 
    \item \small Paul Anak Phillip, N. M. Arsad (2022). Marginalised Students’ Funds of Knowledge in Teaching and Learning Science: A Systematic Literature Review. International Journal of Social Science Research 
    \item \small M. Horton (2022). Funds of Knowledge at San Basilio de Palenque : A path for preserving its identity. International Journal of Multicultural Education 
    \item \small Katherine Espinoza, Idalia Nuñez, E. Degollado (2021). “This is What My Kids See Every Day”: Bilingual Pre-service Teachers Embracing Funds of Knowledge through Border Thinking Pedagogy. Journal of Language, Identity \& Education 20,4 - 17 
    \item \small  (2021). Funds of Knowledge as Pre-College Experiences that Promote Minoritized Students’ Interest, Self-Efficacy Beliefs, and Choice of Majoring in Engineering. Journal of Pre-College Engineering Education Research (J-PEER) 
    \item \small Judith ’t Gilde, M. Volman (2021). Finding and using students’ funds of knowledge and identity in superdiverse primary schools: a collaborative action research project. Cambridge Journal of Education 51,673 - 692 
\end{enumerate}

\subsection{Human-Computer Interaction}

\begin{enumerate}
    \item \small Jeffrey T. Hancock, Mor Naaman, K. Levy (2020). AI-Mediated Communication: Definition, Research Agenda, and Ethical Considerations. J. Comput. Mediat. Commun. 25,89-100 
    \item \small Mina Lee, Percy Liang, Qian Yang (2022). CoAuthor: Designing a Human-AI Collaborative Writing Dataset for Exploring Language Model Capabilities. Proceedings of the 2022 CHI Conference on Human Factors in Computing Systems 
    \item \small Qian Yang, Justin Cranshaw, Saleema Amershi, Shamsi T. Iqbal, J. Teevan (2019). Sketching NLP: A Case Study of Exploring the Right Things To Design with Language Intelligence. Proceedings of the 2019 CHI Conference on Human Factors in Computing Systems 
    \item \small Yewon Kim, Mina Lee, Donghwi Kim, Sung-Ju Lee (2023). Towards Explainable AI Writing Assistants for Non-native English Speakers. ArXiv abs/2304.02625 
    \item \small Elizabeth Clark, Noah A. Smith (2021). Choose Your Own Adventure: Paired Suggestions in Collaborative Writing for Evaluating Story Generation Models. ,3566-3575 
    \item \small A. See, Aneesh S. Pappu, Rohun Saxena, Akhila Yerukola, Christopher D. Manning (2019). Do Massively Pretrained Language Models Make Better Storytellers?. ArXiv abs/1909.10705 
    \item \small Peng Xu, M. Patwary, M. Shoeybi, Raul Puri, Pascale Fung, Anima Anandkumar, Bryan Catanzaro (2020). Controllable Story Generation with External Knowledge Using Large-Scale Language Models. ,2831-2845 
    \item \small Ge Gao, Hao-Chuan Wang, D. Cosley, Susan R. Fussell (2013). Same translation but different experience: the effects of highlighting on machine-translated conversations. Proceedings of the SIGCHI Conference on Human Factors in Computing Systems 
    \item \small Ge Gao, Bin Xu, D. Cosley, Susan R. Fussell (2014). How beliefs about the presence of machine translation impact multilingual collaborations. Proceedings of the 17th ACM conference on Computer supported cooperative work \& social computing 
    \item \small Ge Gao, Bin Xu, David C. Hau, Zheng Yao, D. Cosley, Susan R. Fussell (2015). Two is Better Than One: Improving Multilingual Collaboration by Giving Two Machine Translation Outputs. Proceedings of the 18th ACM Conference on Computer Supported Cooperative Work \& Social Computing 
    \item \small Ge Gao, Naomi Yamashita, Ari Hautasaari, A. Echenique, Susan R. Fussell (2014). Effects of public vs. private automated transcripts on multiparty communication between native and non-native english speakers. Proceedings of the SIGCHI Conference on Human Factors in Computing Systems 
    \item \small Ge Gao, Naomi Yamashita, Ari Hautasaari, Susan R. Fussell (2015). Improving Multilingual Collaboration by Displaying How Non-native Speakers Use Automated Transcripts and Bilingual Dictionaries. Proceedings of the 33rd Annual ACM Conference on Human Factors in Computing Systems 
    \item \small Naomi Yamashita, T. Ishida (2006). Effects of machine translation on collaborative work. ,515-524 
    \item \small Jeffrey T. Hancock, Mor Naaman, K. Levy (2020). AI-Mediated Communication: Definition, Research Agenda, and Ethical Considerations. J. Comput. Mediat. Commun. 25,89-100 
    \item \small Hannah Mieczkowski, Jeffrey T. Hancock, Mor Naaman, Malte F. Jung, Jess Hohenstein (2021). AI-Mediated Communication. Proceedings of the ACM on Human-Computer Interaction 5,1 - 14 
    \item \small Maurice Jakesch, Megan French, Xiao Ma, Jeffrey T. Hancock, Mor Naaman (2019). AI-Mediated Communication: How the Perception that Profile Text was Written by AI Affects Trustworthiness. Proceedings of the 2019 CHI Conference on Human Factors in Computing Systems 
    \item \small Rosanna Fanni, V. Steinkogler, Giulia Zampedri, Joseph Pierson (2022). Enhancing human agency through redress in Artificial Intelligence Systems. Ai \& Society 38,537 - 547 
    \item \small Mor Naaman (2022). “My AI must have been broken”: How AI Stands to Reshape Human Communication. Proceedings of the 16th ACM Conference on Recommender Systems 
    \item \small Zhila Aghajari, Eric P. S. Baumer, Jess Hohenstein, Malte F. Jung, Dominic DiFranzo (2023). Methodological Middle Spaces: Addressing the Need for Methodological Innovation to Achieve Simultaneous Realism, Control, and Scalability in Experimental Studies of AI-Mediated Communication. Proceedings of the ACM on Human-Computer Interaction 7,1 - 28 
    \item \small Nicholas Furlo, Jacob Gleason, Karen Feun, Douglas Zytko (2021). Rethinking Dating Apps as Sexual Consent Apps: A New Use Case for AI-Mediated Communication. Companion Publication of the 2021 Conference on Computer Supported Cooperative Work and Social Computing 
    \item \small Yih-Ling Liu, Anushk Mittal, Diyi Yang, A. Bruckman (2022). Will AI Console Me when I Lose my Pet? Understanding Perceptions of AI-Mediated Email Writing. Proceedings of the 2022 CHI Conference on Human Factors in Computing Systems 
    \item \small Ritika Poddar, R. Sinha, Mor Naaman, Maurice Jakesch (2023). AI Writing Assistants Influence Topic Choice in Self-Presentation. Extended Abstracts of the 2023 CHI Conference on Human Factors in Computing Systems 
    \item \small Tomas Lawton, F. Ibarrola, Dan Ventura, Kazjon Grace (2023). Drawing with Reframer: Emergence and Control in Co-Creative AI. Proceedings of the 28th International Conference on Intelligent User Interfaces 
    \item \small Usman Ahmad Usmani, A. Happonen, J. Watada (2023). Human-Centered Artificial Intelligence: Designing for User Empowerment and Ethical Considerations. 2023 5th International Congress on Human-Computer Interaction, Optimization and Robotic Applications (HORA),01-05 
    \item \small Hyunjin Kang, Chen Lou (2022). AI agency vs. human agency: understanding human-AI interactions on TikTok and their implications for user engagement. J. Comput. Mediat. Commun. 27 
    \item \small Shiye Cao, Chien-Ming Huang (2022). Understanding User Reliance on AI in Assisted Decision-Making. Proceedings of the ACM on Human-Computer Interaction 6,1 - 23 
    \item \small Heidi Toivonen, Francesco Lelli (2020). Agency in Human-Smart Device Relationships: An Exploratory Study.   
    \item \small Sarah Inman, David Ribes (2019). "Beautiful Seams": Strategic Revelations and Concealments. Proceedings of the 2019 CHI Conference on Human Factors in Computing Systems 
\end{enumerate}

\subsection{Immigrant Studies}
\begin{enumerate}
    \item \small Mohammed E. Al-Sofiani, Susan J. Langan, A. Kanaya, N. Kandula, B. Needham, Catherine Kim, Dhananjay Vaidya, S. Golden, K. Gudzune, Clare J. Lee (2020). The relationship of acculturation to cardiovascular disease risk factors among U.S. South Asians: Findings from the MASALA study.. Diabetes research and clinical practice 161,108052 
    \item \small M. Alegría (2009). The challenge of acculturation measures: what are we missing? A commentary on Thomson \& Hoffman-Goetz.. Social science \& medicine 69 7,996-8 
    \item \small Kerstin B Andersson (2019). Digital diasporas: An overview of the research areas of migration and new media through a narrative literature review. Human Technology 
    \item \small A. Balidemaj, M. Small (2019). The effects of ethnic identity and acculturation in mental health of immigrants: A literature review. International Journal of Social Psychiatry 65,643 - 655 
    \item \small Venera Bekteshi, Sung-wan Kang (2020). Contextualizing acculturative stress among Latino immigrants in the United States: a systematic review. Ethnicity \& Health 25,897 - 914 
    \item \small Venera Bekteshi, M. Hook (2015). Contextual Approach to Acculturative Stress Among Latina Immigrants in the U.S.. Journal of Immigrant and Minority Health 17,1401-1411 
    \item \small J. Berry (1992). Acculturation and Adaptation in a New Society. International Migration 30,69-85 
    \item \small J. Berry (1997). Constructing and Expanding a Framework: Opportunities for Developing Acculturation Research. Applied Psychology 46,62-68 
    \item \small S. Bhatia, Anjali Ram (2009). Theorizing identity in transnational and diaspora cultures: A critical approach to acculturation. International Journal of Intercultural Relations 33,140-149 
    \item \small G. Bhattacharya, Susan L. Schoppelrey (2004). Preimmigration Beliefs of Life Success, Postimmigration Experiences, and Acculturative Stress: South Asian Immigrants in the United States. Journal of Immigrant Health 6,83-92 
    \item \small Dania Bilal, I. Bachir (2007). Children's interaction with cross-cultural and multilingual digital libraries. II. Information seeking, success, and affective experience. Inf. Process. Manag. 43,65-80 
    \item \small G. Burnett (2015). Information Worlds and Interpretive Practices: Toward an Integration of Domains. Journal of Information Science Theory and Practice 3,6-16 
    \item \small O. Celenk, Fons J. R. Vijver (2011). Assessment of acculturation : Issues and overview of measures. Online Readings in Psychology and Culture 8,10 
    \item \small R. Cervantes, D. Fisher, A. Padilla, L. Napper (2016). The Hispanic Stress Inventory Version 2: Improving the assessment of acculturation stress.. Psychological assessment 28 5,509-22 
    \item \small Valery Chirkov (2009). Summary of the criticism and of the potential ways to improve acculturation psychology. International Journal of Intercultural Relations 33,177-180 
    \item \small K. Chun, Christine M. L. Kwan, L. Strycker, C. Chesla (2016). Acculturation and bicultural efficacy effects on Chinese American immigrants’ diabetes and health management. Journal of Behavioral Medicine 39,896-907 
    \item \small S. Croucher, E. Kramer (2017). Cultural fusion theory: An alternative to acculturation. Journal of International and Intercultural Communication 10,114 - 97 
    \item \small I. Cuéllar, B. Arnold, R. Maldonado (1995). Acculturation Rating Scale for Mexican Americans-II: A Revision of the Original ARSMA Scale. Hispanic Journal of Behavioral Sciences 17,275 - 304 
    \item \small Zita Dixon, Melissa L. Bessaha, Margaret A. Post (2018). Beyond the Ballot: Immigrant Integration Through Civic Engagement and Advocacy. Race and Social Problems 10,366 - 375 
    \item \small R. Dekker, G. Engbersen, M. Faber (2016). The Use of Online Media in Migration Networks. Population Space and Place 22,539-551 
    \item \small K. Fisher, J. Durrance, M. Hinton (2004). Information grounds and the use of need-based services by immigrants in Queens, New York: A context-based, outcome evaluation approach. J. Assoc. Inf. Sci. Technol. 55,754-766 
    \item \small K. Fisher, Charles M. Naumer (2006). Information Grounds: Theoretical Basis and Empirical Findings on Information Flow in Social Settings.  ,93-111 
    \item \small N. Harney (2013). Precarity, Affect and Problem Solving with Mobile Phones by Asylum Seekers, Refugees and Migrants in Naples, Italy. Journal of Refugee Studies 26,541-557 
    \item \small A. Inkeles (1969). Making Men Modern: On the Causes and Consequences of Individual Change in Six Developing Countries. American Journal of Sociology 75,208 - 225 
    \item \small P. Jaeger, G. Burnett (2010). Information Worlds: Behavior, Technology, and Social Context in the Age of the Internet.   
    \item \small Y. Y. Kim (1976). A Causal Model of Communication Patterns of Foreign Immigrants in the Process of Acculturation..   
    \item \small E. Klonoff, H. Landrine (2000). Revising and Improving the African American Acculturation Scale. Journal of Black Psychology 26,235 - 261 
    \item \small E. Kramer (2012). 7 Dimensional Accrual and Dissociation an Introduction.   
    \item \small S. Lee (2013). “Flow” in Art Therapy: Empowering Immigrant Children With Adjustment Difficulties. Art Therapy 30,56 - 63 
    \item \small Jiwon R Lee, N. Maruthur, H. Yeh (2020). Nativity and prevalence of cardiometabolic diseases among U.S. Asian immigrants.. Journal of diabetes and its complications,107679 
    \item \small G. Marín, R. J. Gamba (1996). A New Measurement of Acculturation for Hispanics: The Bidimensional Acculturation Scale for Hispanics (BAS). Hispanic Journal of Behavioral Sciences 18,297 - 316 
    \item \small A. Meca, S. Schwartz, C. Martinez, Heather McClure (2018). Longitudinal effects of acculturation and enculturation on mental health: Does the measure of matter?. Development and Psychopathology 30,1849 - 1866 
    \item \small M. I. Mercado (1997). Multicultural realities in the information age: questions of relevance for public library resource allocation and adequacy. The Bottom Line: Managing Library Finances 10,119-122 
    \item \small Jane M. Murphy (1973). Sociocultural Change and Psychiatric Disorder Among Rural Yorubas in Nigeria. Ethos 1,239-262 
    \item \small N. Nagaraj, Amita N. Vyas, K. McDonnell, L. DiPietro (2018). Understanding Health, Violence, and Acculturation Among South Asian Women in the US. Journal of Community Health 43,543-551 
    \item \small M. Nedelcu (2012). Migrants' New Transnational Habitus: Rethinking Migration Through a Cosmopolitan Lens in the Digital Age. Journal of Ethnic and Migration Studies 38,1339 - 1356 
    \item \small C. Negy, S. Schwartz, A. Reig-Ferrer (2009). Violated expectations and acculturative stress among U.S. Hispanic immigrants.. Cultural diversity \& ethnic minority psychology 15 3,255-64 
    \item \small Thomas G. Plante (2023). Work is religion and religion is work in Silicon Valley. Politics, Religion \& Ideology 25,309 - 311 
    \item \small L. H. Rogler, D. Cortés, R. Malgady (1991). Acculturation and mental health status among Hispanics. Convergence and new directions for research.. The American psychologist 46 6,585-97 
    \item \small F. Rudmin (2009). Constructs, measurements and models of acculturation and acculturative stress. International Journal of Intercultural Relations 33,106-123 
    \item \small Maria D Thomson, L. Hoffman‐Goetz (2009). Defining and measuring acculturation: a systematic review of public health studies with Hispanic populations in the United States.. Social science \& medicine 69 7,983-91 
    \item \small Peter Weinreich (2009). ‘Enculturation’, not ‘acculturation’: Conceptualising and assessing identity processes in migrant communities. International Journal of Intercultural Relations 33,124-139 
    \item \small Xuanxuan Zhu, Jihong Liu, M. Sevoyan, R. Pate (2021). Acculturation and leisure-time physical activity among Asian American adults in the United States. Ethnicity \& Health 27,1900 - 1914 
\end{enumerate}

\subsection{Ethnic-Racial Studies}

\begin{enumerate}
    \item \small Deborah Rivas‐Drake, Eleanor K. Seaton, C. Markstrom, S. Quintana, M. Syed, Richard M. Lee, S. Schwartz, A. Umaña‐Taylor, S. French, T. Yip (2014). Ethnic and racial identity in adolescence: implications for psychosocial, academic, and health outcomes.. Child development 85 1,40-57 
    \item \small Ronald Krate, Gloria Leventhal, B. Silverstein (1974). Self-Perceived Transformation of Negro-to-Black Identity. Psychological Reports 35,1071 - 1075 
    \item \small R. Roberts, J. Phinney, L. Mâsse, Y. R. Chen, C. R. Roberts, A. Romero (1999). The Structure of Ethnic Identity of Young Adolescents from Diverse Ethnocultural Groups. The Journal of Early Adolescence 19,301 - 322 
    \item \small S. Quintana (1998). Children's developmental understanding of ethnicity and race. Applied \& Preventive Psychology 7,27-45 
    \item \small Melinda A Gonzales-Backen (2013). An Application of Ecological Theory to Ethnic Identity Formation Among Biethnic Adolescents. Family Relations 62,92-108 
    \item \small Sara Douglass, A. Umaña‐Taylor (2015). A Brief Form of the Ethnic Identity Scale: Development and Empirical Validation. Identity 15,48 - 65 
    \item \small A. Umaña‐Taylor, M. Fine (2001). Methodological Implications of Grouping Latino Adolescents into One Collective Ethnic Group. Hispanic Journal of Behavioral Sciences 23,347 - 362 
    \item \small A. Umaña‐Taylor, Ani Yazedjian, M. Bámaca-Gómez (2004). Developing the Ethnic Identity Scale Using Eriksonian and Social Identity Perspectives. Identity 4,38 - 9 
    \item \small K. Bush, A. Supple, Sheryl Beaty Lash (2004). Mexican Adolescents' Perceptions of Parental Behaviors and Authority as Predictors of Their Self-Esteem and Sense of Familism. Marriage \& Family Review 36,35 - 65 
    \item \small Gina M. Shkodriani, J. Gibbons (1995). Individualism and Collectivism among University Students in Mexico and the United States. Journal of Social Psychology 135,765-772 
    \item \small A. Supple, S. Ghazarian, James M. Frabutt, S. Plunkett, Tovah Sands (2006). Contextual influences on Latino adolescent ethnic identity and academic outcomes.. Child development 77 5,1427-33 
    \item \small A. Umaña‐Taylor, Ani Yazedjian, M. Bámaca-Gómez (2004). Developing the Ethnic Identity Scale Using Eriksonian and Social Identity Perspectives. Identity 4,38 - 9 
    \item \small Katharina Schmid, Nicole Tausch, M. Hewstone (2009). The Social Psychology of intergroup relations.   
    \item \small H. Tajfel (1970). Aspects of national and ethnic loyalty. Social Science Information 9,119 - 144 
    \item \small J. Phinney (1989). Stages of Ethnic Identity Development in Minority Group Adolescents. The Journal of Early Adolescence 9,34 - 49 
    \item \small Jan E. Stets, P. Burke (2000). Identity theory and social identity theory. Social Psychology Quarterly 63,224-237 
    \item \small H. Tajfel, G. Jahoda, C. Nemeth, Y. Rim, N. Johnson (1972). The Devaluation by Children of their Own National and Ethnic Group: Two Case Studies. The British journal of social and clinical psychology 11,235-243 
    \item \small A. Umaña‐Taylor, Edna C. Alfaro, M. Bámaca, Amy B. Guimond (2009). The central role of familial ethnic socialization in latino adolescents' cultural orientation. Journal of Marriage and Family 71,46-60 
    \item \small A. Umaña‐Taylor, Ruchi Bhanot, Nana Shin (2006). Ethnic Identity Formation During Adolescence. Journal of Family Issues 27,390 - 414 
    \item \small A. Umaña‐Taylor, M. Diversi, M. Fine (2002). Ethnic Identity and Self-Esteem of Latino Adolescents. Journal of Adolescent Research 17,303 - 327 
    \item \small A. Umaña‐Taylor, M. Fine (2004). Examining Ethnic Identity among Mexican-Origin Adolescents Living in the United States. Hispanic Journal of Behavioral Sciences 26,36 - 59 
\end{enumerate}

\subsection{Professional Identity}

\begin{enumerate}
    \item \small Lynne Siemens, Yin Liu, Jeff Smith (2014). Mapping Disciplinary Differences and Equity of Academic Control to Create a Space for Collaboration.. Canadian Journal of Higher Education 44,49-67 
    \item \small L. Lingard, K. Garwood, Catherine F. Schryer, M. Spafford (2003). A certain art of uncertainty: case presentation and the development of professional identity.. Social science \& medicine 56 3,603-16 
    \item \small J. Dart, S. Ash, L. McCall, C. Rees (2022). 'We're our own worst enemies': A qualitative exploration of sociocultural factors in dietetic education influencing student-dietitian transitions.. Journal of the Academy of Nutrition and Dietetics 
    \item \small Sharon Fraser, K. Beswick, Margaret Penson, A. Seen, R. Whannell (2019). Cross Faculty Collaboration in the Development of an Integrated Mathematics and Science Initial Teacher Education Program. Australian Journal of Teacher Education 
    \item \small Seongsook Choi, Keith Richards (2017). The dynamics of identity struggle in interdisciplinary meetings in higher education.  ,165-184 
    \item \small J. Thistlethwaite, Koshila Kumar, C. Roberts (2016). Becoming Interprofessional: Professional Identity Formation in the Health Professions.  ,140-154 
    \item \small Lars Geschwind, G. Melin (2016). Stronger disciplinary identities in multidisciplinary research schools. Studies in Continuing Education 38,16 - 28 
    \item \small L. Allen, Kathleen Quinlivan, C. Aspin, Fida Sanjakdar, A. Bromdal, M. Rasmussen (2014). Meeting at the crossroads: re-conceptualising difference in research teams. Qualitative Research Journal 14,119-133 
    \item \small C. Rees, L. Monrouxe (2018). Who are you and who do you want to be? Key considerations in developing professional identities in medicine. Medical Journal of Australia 209 
    \item \small S. Cruess, R. Cruess, Y. Steinert (2019). Supporting the development of a professional identity: General principles. Medical Teacher 41,641 - 649 
    \item \small Sally G. Warmington, G. McColl (2016). Medical student stories of participation in patient care-related activities: the construction of relational identity. Advances in Health Sciences Education 22,147 - 163 
    \item \small A. O’Cathain, E. Murphy, J. Nicholl (2008). Multidisciplinary, Interdisciplinary, or Dysfunctional? Team Working in Mixed-Methods Research. Qualitative Health Research 18,1574 - 1585 
    \item \small E. Schipper, N. Dubash, Y. Mulugetta (2021). Climate change research and the search for solutions: rethinking interdisciplinarity. Climatic Change 168 
    \item \small W. Bruin, W. Bruin, M. G. Morgan (2019). Reflections on an interdisciplinary collaboration to inform public understanding of climate change, mitigation, and impacts. Proceedings of the National Academy of Sciences 116,7676 - 7683 
    \item \small Susan K. Gardner (2013). Paradigmatic differences, power, and status: a qualitative investigation of faculty in one interdisciplinary research collaboration on sustainability science. Sustainability Science 8,241-252 
    \item \small A. Brew (2008). Disciplinary and interdisciplinary affiliations of experienced researchers. Higher Education 56,423-438 
    \item \small Valérie Tartas, Nathalie Muller Mirza (2007). Rethinking Collaborative Learning Through Participation in an Interdisciplinary Research Project: Tensions and Negotiations as Key Points in Knowledge Production. Integrative Psychological and Behavioral Science 41,154-168 
    \item \small B. Choi, Anita W. P. Pak (2007). Multidisciplinarity, interdisciplinarity, and transdisciplinarity in health research, services, education and policy: 2. Promotors, barriers, and strategies of enhancement.. Clinical and investigative medicine. Medecine clinique et experimentale 30 6,E224-32 
    \item \small L. Lingard, Catherine F. Schryer, M. Spafford, S. Campbell (2007). Negotiating the politics of identity in an interdisciplinary research team. Qualitative Research 7,501 - 519 
    \item \small Xin Ming, M. MacLeod, Jan T. van der Veen (2023). Construction and enactment of interdisciplinarity: A grounded theory case study in Liberal Arts and Sciences education. Learning, Culture and Social Interaction 
    \item \small Daniel Thorpe (2023). Dependent or not? From a daily practice of Earth observation research in the Global South to promoting adequate developmental spaces in science and technology studies. Geographica Helvetica 
    \item \small L. Livingston (2022). Partnerships in pandemics: tracing power relations in community engaged scholarship in food systems during COVID-19. Agriculture and Human Values 40,217 - 229 
    \item \small  (2022). OUP accepted manuscript. Science and Public Policy 
    \item \small Brittany I. Davidson, Katiuska Mara Ferrer Portillo, Marcli Wac, C. McWilliams, C. Bourdeaux, I. Craddock (2021). Requirements for a Bespoke Intensive Care Unit Dashboard in Response to the COVID-19 Pandemic: Semistructured Interview Study. JMIR Human Factors 9 
    \item \small B. E. Arias López, Christine Andrä, Berit Bliesemann de Guevara (2021). Reflexivity in research teams through narrative practice and textile-making. Qualitative Research 23,306 - 312 
    \item \small J. Leigh, N. Brown (2021). Researcher experiences in practice-based interdisciplinary research. Research Evaluation 
    \item \small Emily J. Noonan, L. Weingartner, Ryan. M. Combs, C. Bohnert, M. Ann Shaw, Susan Sawning (2020). Perspectives of Transgender and Genderqueer Standardized Patients. Teaching and Learning in Medicine 33,116 - 128 
    \item \small M. Roura (2020). The Social Ecology of Power in Participatory Health Research. Qualitative Health Research 31,778 - 788 
    \item \small C. Barry, N. Britten, N. Barber, C. Bradley, Fiona A. Stevenson (1999). Using Reflexivity to Optimize Teamwork in Qualitative Research. Qualitative Health Research 9,26 - 44 
    \item \small L. Bracken, E. Oughton (2006). ‘What do you mean?’ The importance of language in developing interdisciplinary research. Transactions of the Institute of British Geographers 31,371-382 
    \item \small Robin H. Rogers-Dillon (2005). Hierarchical qualitative research teams: refining the methodology. Qualitative Research 5,437 - 454 
    \item \small L. Lingard, Catherine F. Schryer, Kim C. Garwood, M. Spafford (2003). ‘Talking the talk’: school and workplace genre tension in clerkship case presentations. Medical Education 37 
    \item \small Catherine F. Schryer, Innis (2011). 2 INVESTIGATING TEXTS IN THEIR SOCIAL CONTEXTS: THE PROMISE AND PERIL OF RHETORICAL GENRE STUDIES.  
    \item \small Claire Massey, F. Alpass, R. Flett, Kate Lewis, S. Morriss, F. Sligo (2006). Crossing fields: the case of a multi-disciplinary research team. Qualitative Research 6,131 - 147 
    \item \small E. Wenger (1998). Communities of Practice: Learning, Meaning, and Identity.   
    \item \small Trena M Paulus, M. Woodside, M. Ziegler (2010). “I Tell You, It’s a Journey, Isn’t It?” Understanding Collaborative Meaning Making in Qualitative Research. Qualitative Inquiry 16,852 - 862 
    \item \small M. Spafford, Catherine F. Schryer, S. Creutz (2009). Balancing patient care and student education: Learning to deliver bad news in an optometry teaching clinic. Advances in Health Sciences Education 14,233-250 
    \item \small Lynne Siemens, Yin Liu, Jeff Smith (2014). Mapping Disciplinary Differences and Equity of Academic Control to Create a Space for Collaboration.. Canadian Journal of Higher Education 44,49-67 
    \item \small L. Lingard, K. Garwood, Catherine F. Schryer, M. Spafford (2003). A certain art of uncertainty: case presentation and the development of professional identity.. Social science \& medicine 56 3,603-16 
    \item \small J. Dart, S. Ash, L. McCall, C. Rees (2022). 'We're our own worst enemies': A qualitative exploration of sociocultural factors in dietetic education influencing student-dietitian transitions.. Journal of the Academy of Nutrition and Dietetics 
    \item \small Sharon Fraser, K. Beswick, Margaret Penson, A. Seen, R. Whannell (2019). Cross Faculty Collaboration in the Development of an Integrated Mathematics and Science Initial Teacher Education Program. Australian Journal of Teacher Education 
    \item \small Seongsook Choi, Keith Richards (2017). The dynamics of identity struggle in interdisciplinary meetings in higher education.  ,165-184 
    \item \small J. Thistlethwaite, Koshila Kumar, C. Roberts (2016). Becoming Interprofessional: Professional Identity Formation in the Health Professions.  ,140-154 
    \item \small Lars Geschwind, G. Melin (2016). Stronger disciplinary identities in multidisciplinary research schools. Studies in Continuing Education 38,16 - 28 
    \item \small L. Allen, Kathleen Quinlivan, C. Aspin, Fida Sanjakdar, A. Bromdal, M. Rasmussen (2014). Meeting at the crossroads: re-conceptualising difference in research teams. Qualitative Research Journal 14,119-133 
    \item \small C. Rees, L. Monrouxe (2018). Who are you and who do you want to be? Key considerations in developing professional identities in medicine. Medical Journal of Australia 209 
    \item \small S. Cruess, R. Cruess, Y. Steinert (2019). Supporting the development of a professional identity: General principles. Medical Teacher 41,641 - 649 
    \item \small Sally G. Warmington, G. McColl (2016). Medical student stories of participation in patient care-related activities: the construction of relational identity. Advances in Health Sciences Education 22,147 - 163 
    \item \small A. O’Cathain, E. Murphy, J. Nicholl (2008). Multidisciplinary, Interdisciplinary, or Dysfunctional? Team Working in Mixed-Methods Research. Qualitative Health Research 18,1574 - 1585 
    \item \small E. Schipper, N. Dubash, Y. Mulugetta (2021). Climate change research and the search for solutions: rethinking interdisciplinarity. Climatic Change 168 
    \item \small W. Bruin, W. Bruin, M. G. Morgan (2019). Reflections on an interdisciplinary collaboration to inform public understanding of climate change, mitigation, and impacts. Proceedings of the National Academy of Sciences 116,7676 - 7683 
    \item \small Susan K. Gardner (2013). Paradigmatic differences, power, and status: a qualitative investigation of faculty in one interdisciplinary research collaboration on sustainability science. Sustainability Science 8,241-252 
    \item \small A. Brew (2008). Disciplinary and interdisciplinary affiliations of experienced researchers. Higher Education 56,423-438 
    \item \small Valérie Tartas, Nathalie Muller Mirza (2007). Rethinking Collaborative Learning Through Participation in an Interdisciplinary Research Project: Tensions and Negotiations as Key Points in Knowledge Production. Integrative Psychological and Behavioral Science 41,154-168 
    \item \small B. Choi, Anita W. P. Pak (2007). Multidisciplinarity, interdisciplinarity, and transdisciplinarity in health research, services, education and policy: 2. Promotors, barriers, and strategies of enhancement.. Clinical and investigative medicine. Medecine clinique et experimentale 30 6,E224-32 
\end{enumerate}